\documentclass{ieeeoj}
\usepackage{cite}
\usepackage{amsmath,amsfonts,amssymb,amsthm}
\usepackage{algorithmic}
\usepackage{algorithm}
\usepackage{array}
\usepackage[caption=false]{subfig}
\usepackage{textcomp}
\usepackage{stfloats}
\usepackage{xurl}
\usepackage{graphicx,color}
\usepackage{bm}
\usepackage{bbm}
\usepackage{orcidlink}
\let\labelindent\relax
\usepackage{enumitem}
\usepackage{placeins}
\usepackage{mathtools}
\usepackage{capt-of}
\usepackage{cuted}
\usepackage{caption}
\usepackage{float}
\hypersetup{hidelinks}
\makeatletter
\def\Hy@Warning#1{}
\let\OJlogo\@empty
\let\@revised\@empty
\makeatother

\newcommand{\R}{\mathbb{R}}
\newcommand{\E}{\mathbb{E}}

\newcommand{\norm}[1]{\left\lVert #1 \right\rVert}
\newcommand{\abs}[1]{\left\lvert #1 \right\rvert}

\newcommand{\diag}{\mathrm{diag}}

\newcommand{\Proj}{\mathrm{Proj}}

\newcommand{\sep}{, }

\theoremstyle{plain}
\newtheorem{theorem}{Theorem}

\theoremstyle{definition}
\newtheorem{assumption}{Assumption}

\begin{document}

\renewcommand{\topfraction}{.85}
\renewcommand{\bottomfraction}{.7}
\renewcommand{\textfraction}{.15}
\renewcommand{\floatpagefraction}{.66}
\renewcommand{\dbltopfraction}{.66}
\renewcommand{\dblfloatpagefraction}{.66}
\setcounter{topnumber}{9}
\setcounter{bottomnumber}{9}
\setcounter{totalnumber}{20}
\setcounter{dbltopnumber}{9}

\title{A Distributionally Robust Reinforcement Learning Framework for Constrained Urban EV Dispatch}

\author{An Nguyen\orcidlink{0009-0001-3971-7557}\textsuperscript{1},
Hoang Nguyen\orcidlink{0009-0001-3771-9305}\textsuperscript{1},
Phuong Le\orcidlink{0009-0009-5095-8763}\textsuperscript{1},
Hung Pham\orcidlink{0009-0006-5209-9446}\textsuperscript{1},
Cuong Do\orcidlink{0000-0003-4785-4524}\textsuperscript{1,2},
and Laurent El Ghaoui\orcidlink{0000-0002-0499-5610}\textsuperscript{1,2}}

\affil{College of Engineering and Computer Science, VinUniversity, Hanoi, Vietnam}
\affil{Center for Environmental Intelligence, VinUniversity, Hanoi, Vietnam}
\corresp{CORRESPONDING AUTHORS:  An Nguyen (e-mail: 23an.nt@vinuni.edu.vn), Cuong Do (e-mail: cuong.dd@vinuni.edu.vn)}
\markboth{A Distributionally Robust Reinforcement Learning Framework for Constrained Urban EV Dispatch}{Nguyen \textit{et al.}}

\begin{abstract}
We study city-scale control of electric-vehicle (EV) ride-hailing fleets where dispatch, repositioning, and charging decisions must respect charger and feeder limits under uncertain, spatially correlated demand and travel times. We formulate the problem as a hex-grid semi-Markov decision process (semi-MDP) with mixed actions---discrete (serve, reposition, charge) and continuous (charging power)---and variable action durations. To guarantee physical feasibility at both training and deployment, we learn over \emph{intentions} produced by a masked, temperature-annealed policy. These intentions are projected, at every decision step, through a time-limited rolling mixed-integer linear program (MILP) that strictly enforces state-of-charge, port, and feeder constraints. To mitigate distributional shifts, we optimize a Soft Actor--Critic (SAC) agent against a \emph{true} Wasserstein-1 ambiguity set, employing a graph-aligned Mahalanobis ground metric to capture spatial correlations. The robust backup utilizes the Kantorovich--Rubinstein dual with a projected subgradient inner loop and a primal--dual risk budget update. Our architecture combines a two-layer Graph Convolutional Network (GCN) encoder, twin critics, and a value network driving the adversary. Experiments on a large-scale EV fleet simulator built from NYC taxi data show that PD--RSAC achieves the highest net profit, reaching \$1.22M, compared with \$0.58M--\$0.70M for strong heuristic, single-agent RL, and multi-agent RL baselines, including Greedy, SAC, MAPPO, and MADDPG, while maintaining zero feeder-limit violations.
\end{abstract}

\begin{IEEEkeywords}
Reinforcement learning \sep Distributionally robust optimization \sep Graph neural networks \sep Constrained decision making \sep Electric vehicle fleet management
\end{IEEEkeywords}

\maketitle

\section{Introduction}
Electric-vehicle (EV) ride-hailing fleets are increasingly crucial for urban mobility, requiring operators to continuously assign vehicles to requests, reposition idle units, and schedule charging. These decisions must strictly adhere to state-of-charge (SoC) safety limits, charger-port capacities, and power caps, all while maintaining high service quality. However, demand and travel times vary stochastically over space and time, exhibiting correlated and nonstationary patterns. Consequently, fleet control represents a challenging constrained, sequential decision-making problem under uncertainty.

\subsection{Optimization vs.~Learning.} 
Early efforts in fleet coordination primarily relied on optimization-based techniques. For instance, \cite{Wei2021} formulated a bilevel optimization model for multimodal transit, while \cite{Cao2021} addressed user fairness via integer programming. Similarly, \cite{Miao2015}  proposed receding-horizon control (RHC) for dispatching based on real-time sensing. While these approaches explicitly encode operational constraints, they rely heavily on accurate forecasts and suffer from scalability issues as system size increases. Conversely, Reinforcement Learning (RL) has been adopted to improve scalability and efficiency. Recent works optimize dispatching \cite{Liu2022, Qin2020, Yue2024, Wang2025}, vehicle repositioning \cite{Jiao2021, Li2022}, and dynamic pricing \cite{Huang2022, Jin2025} using Deep Q-Networks or Actor--Critic methods. A closely related study by ~\cite{Tian2025} combines MAPPO with binary linear programming for Shared Autonomous EV scheduling. Graph-based multi-agent reinforcement learning has also been explored for structured coordination; for example, Hu et al.~\cite{Hu2024} propose a decentralized graph-based MARL framework using reward machines. Despite their empirical success, standard RL methods typically fail to guarantee \emph{hard constraints} (e.g., grid limits) and remain brittle to distributional shifts between training and deployment.

\subsection{The Challenge of Feasibility.} 
Ensuring safety in RL is a longstanding challenge. Approaches like Constrained Policy Optimization \cite{Achiam2017} or Lagrangian relaxation satisfy constraints only in expectation, which is insufficient for grid-constrained operations where a single violation can trip a feeder. While differentiable optimization layers like OptNet \cite{Amos2017} and CvxpyLayers \cite{Agrawal2019} can project actions into feasible sets, they are limited to convex problems. Related efforts on safe and constrained reinforcement learning include the control-barrier-function-based framework of Liu et al.~\cite{Liu2023}, which studies affine nonlinear systems with state constraints and input saturation. Our domain involves discrete decisions (dispatch modes) and non-convex dynamics, necessitating a mixed-integer projection. Unlike prior works using heuristic repairs, we propose embedding a time-limited rolling MILP directly into the decision loop to guarantee physical feasibility.

\subsection{Decision Making Under Uncertainty.} Furthermore, standard RL assumes identical training and testing dynamics. To address performance degradation under distributional shifts, robust MDPs optimize worst-case performance. Early works \cite{Iyengar2005, Nilim2005} used KL-divergence or rectangular ambiguity sets, often leading to overly conservative policies. More recently, Wasserstein-based ambiguity sets have been incorporated into robust MDPs and reinforcement learning \cite{GrandClement2021, Kordabad2022}, enabling principled robustness against distributional perturbations with provable guarantees. Foundational studies in \textit{Distributionally Robust Optimization (DRO)} demonstrate that Wasserstein balls can effectively capture geometric perturbations in data distributions \cite{Esfahani2018} and establish a close connection between DRO and adversarial training \cite{Sinha2018}. However, standard Wasserstein approaches mainly focus on unconstrained settings with continuous state spaces and typically rely on Euclidean ground metrics ($L_2$ distance), which ignore the underlying graph topology and fail to capture network-induced spatial interactions in transportation systems. In this work, we introduce a graph-aligned ground metric and integrate Wasserstein DRO with mixed-integer feasibility guarantees for spatially structured transportation grids.

\subsection{Contributions.} 
We develop a planning-aware, distributionally robust RL framework for controlling EV fleets on a hexagonal grid. Instead of executing actions directly, our policy outputs \emph{intentions} that are projected through a rolling MILP. To address uncertainty, we optimize a Soft Actor--Critic (SAC) \cite{Haarnoja2018} agent against a Wasserstein-1 ambiguity set equipped with a graph-aligned Mahalanobis metric. Our contributions are threefold:
\begin{enumerate}[label=(\roman*),itemsep=0pt,topsep=2pt,parsep=0pt,partopsep=0pt]
    \item A semi-MDP formulation with duration-aware backups;
    \item A feasible-action architecture embedding a rolling MILP projection; and
    \item A graph-aligned Wasserstein-1 robust RL method that mitigates spatially correlated uncertainty.
\end{enumerate}

\section{System Model and WDRO Objective}
\label{sec:system}

\begin{figure}
    \centering \includegraphics[width=3.5in]{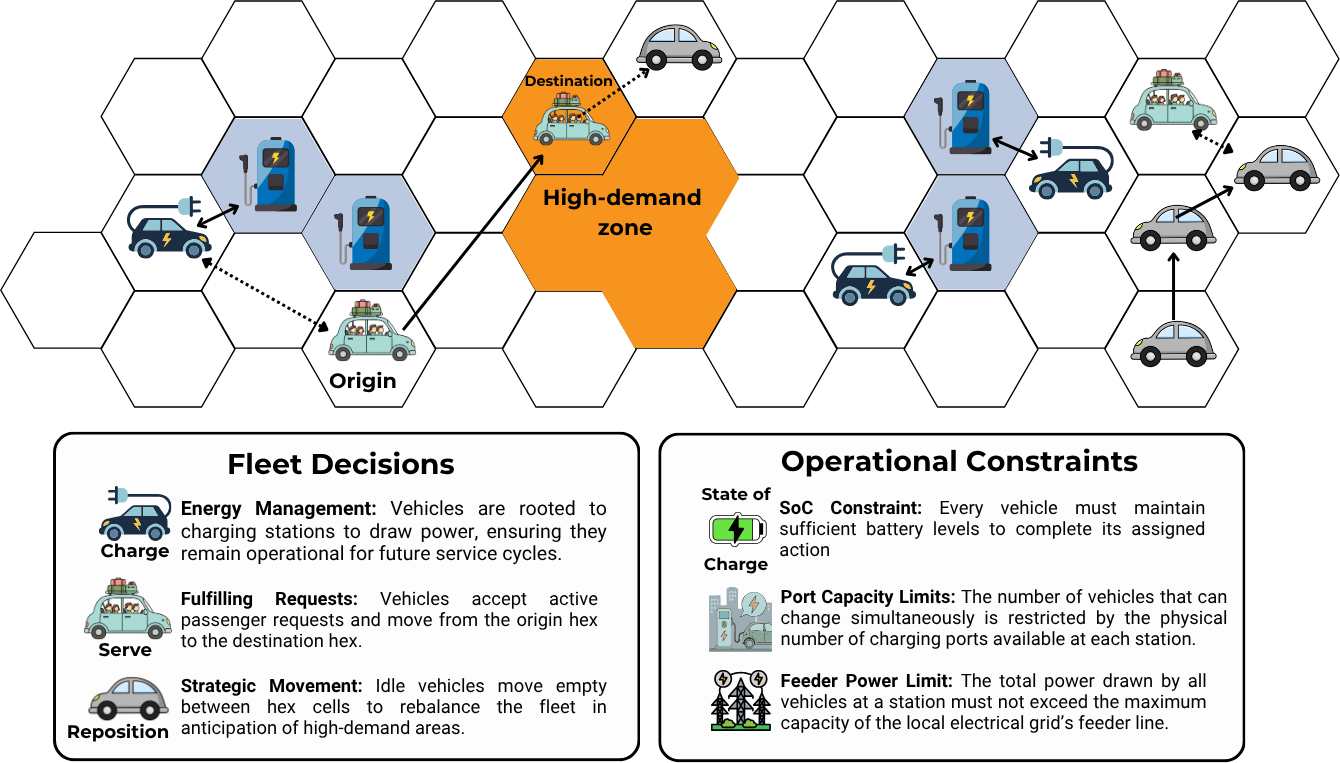}
    \caption{Illustration of the city-scale EV ride-hailing problem on a hexagonal grid.}
    \label{fig:problem_setting}
\end{figure}

Figure~\ref{fig:problem_setting} illustrates the city-scale EV ride-hailing setting considered in this work, including the hex-grid partition, the main fleet decisions, and the key operational constraints. We consider a discrete-time horizon $t\in\{0,1,\dots,T-1\}$ with step size $\Delta t>0$. The service region is discretized into a hexagonal grid $\mathcal H=\{1,\dots,m\}$, where each hex $h$ has a set of neighbors $\mathcal N(h)$.
The fleet comprises $N$ electric vehicles $\mathcal V=\{1,\dots,N\}$. Charging infrastructure is distributed across a subset of hexes $\mathcal S\subseteq\mathcal H$. Each station $s \in \mathcal S$ is characterized by:
\begin{itemize}
    \item $C_s^{\text{port}}$: total number of physical charging plugs (ports) available.
    \item $P_s^{\max}$: maximum charging power per vehicle at station $s$.
    \item $P_{\text{feed}}^{\max}$: total power capacity of the local grid feeder supplying the station set.
\end{itemize}
Vehicle $i$ is modeled by its current battery energy (SoC) $E_{i,t}\in[0,E_i^{\max}]$, charging efficiency $\eta_c\in(0,1]$, and energy consumption rate $\eta_{\text{drv}}>0$ per unit distance. For $Q\succ 0$, we use the shorthand $\norm{x}_Q := \sqrt{x^\top Q x}$.

\subsection{Hex discretization and exogenous scenarios}
At each step $t$, the system environment is driven by an exogenous scenario $\xi_t = (D_t,T_t)$. Here, $D_t\in\R_+^{m\times m}$ represents the Origin-Destination (OD) demand intensity, where $D_t(h_1,h_2)$ is the expected number of requests from hex $h_1$ to $h_2$. $T_t\in\mathbb N^{m\times m}$ denotes the discrete travel time field in multiples of $\Delta t$, with $T_t(h_1,h_2)\ge 1$. These parameters are stochastic and spatially correlated.

\subsection{Aggregated system state}


To enable scalable learning, we adopt an aggregate state representation $s_t$ that captures fleet dynamics at a hex-level granularity rather than tracking individual vehicles:
\begin{equation}
\label{eq:state}
s_t=\Big(X_t^{\text{fleet}},\ X_t^{\text{demand}},\ X_t^{\text{infra}},\ t\Big).
\end{equation}
Specifically:
\begin{itemize}
    \item $X_t^{\text{fleet}}=\{n^{\text{idle}}_{h,t},n^{\text{busy}}_{h,t},\bar E^{(b)}_{h,t}\}_{h\in\mathcal H}$ summarizes the counts of idle/busy vehicles and the average SoC of busy vehicles in each zone.
    \item $X_t^{\text{demand}}=\{D_t(h,\cdot),D_t(\cdot,h)\}_{h\in\mathcal H}$ summarizes the demand outflow and inflow for each hex.
    \item $X_t^{\text{infra}}=\{u_{s,t},q_{s,t},p^{\text{elec}}_{s,t}\}_{s\in\mathcal S}\cup\{P_{\text{feed}}^{\max}\}$ captures station utilization $u$, queue length $q$, dynamic electricity price $p^{\text{elec}}$, and the feeder limit $P_{\text{feed}}^{\max}$.
\end{itemize}

\subsection{Scheduling objective under distributional uncertainty}

Given a stationary policy $\pi$, the (episode) discounted return along a scenario sequence $\xi=(\xi_0,\dots,\xi_{T-1})$ is
\[
R(\xi;\pi)=\E\!\left[\sum_{t=0}^{T-1}\gamma^t r_t\right],
\]
where $r_t$ is the reward at decision epoch $t$ (defined in Sec.~\ref{sec:semi_mdp}), and $0<\gamma<1$ is a discount factor.

Let $\hat P$ denote the empirical distribution of scenarios obtained from historical data. To hedge against distributional shifts, we optimize a distributionally robust objective over a Wasserstein-1 ball of radius $\rho>0$ (under a metric $d_Q$ defined later in Sec.~\ref{sec:mag}):
\begin{equation}
\label{eq:wdro_outer}
\max_{\pi}\ \ \inf_{P:\,W^{d_Q}_1(P,\hat P)\le \rho}\ \ \E_{\xi\sim P}\!\left[R(\xi;\pi)\right].
\end{equation}
The robust backup used in our semi-MDP implementation is a sample-based approximation of~\eqref{eq:wdro_outer}, given in Sec.~\ref{sec:mag}.

\section{Hex Semi-MDP Formulation}
\label{sec:semi_mdp}

\noindent Since transportation tasks (trips) and charging sessions have variable durations that may span multiple time steps $\Delta t$, we model the control problem as a Semi-Markov Decision Process (semi-MDP).

\subsection{Per-vehicle actions and SoC Guard}
For vehicle $i$ at hex $h_{i,t}$ with SoC $E_{i,t}$, the set of valid candidate actions $\mathcal C_{i,t}$ includes:
\begin{equation}
\label{eq:candidate}
\mathcal C_{i,t}=\underbrace{\mathcal O_{i,t}^{\text{serve}}}_{\text{orders}}\ \cup\ \underbrace{\mathcal N(h_{i,t})}_{\text{reposition}}\ \cup\ \underbrace{\mathcal S_{i,t}^{\text{chg}}}_{\text{stations}}.
\end{equation}
However, before an order $o=(h_o^{\text{pu}}\!\to h_o^{\text{do}})$ is considered a valid candidate, it must pass a State-of-Charge (SoC) guard ensuring the vehicle has enough energy to reach the pickup point, complete the trip, and maintain a safety buffer $E_i^{\min}$:
\begin{equation}
\label{eq:soc_guard}
E_{i,t}\ \ge\ \eta_{\text{drv}}\Big(e^{\text{pk}}(h_{i,t},h_o^{\text{pu}})+e^{\text{trip}}(h_o^{\text{pu}},h_o^{\text{do}})\Big)+E_i^{\min}.
\end{equation}
Here $e^{\text{pk}}(h_1,h_2)$ and $e^{\text{trip}}(h_1,h_2)$ denote the energy required for pickup and trip legs, respectively. If this condition fails, the order is masked out from the vehicle's action space.

The per-vehicle action $a_{i,t}$ is a mixed tuple of discrete assignments and continuous power:
\begin{equation}
\label{eq:action_tuple}
a_{i,t}=\big(\{x_{io}\}_{o\in\mathcal O_{i,t}^{\text{serve}}},\ \{y_{ig}\}_{g\in\mathcal N(h_{i,t})},\ \{z_{is}\}_{s\in\mathcal S_{i,t}^{\text{chg}}},\ p_{i,t}\big),
\end{equation}
subject to the constraints:
\begin{equation}
\label{eq:action_constraints}
\begin{split}
& x_{io},y_{ig},z_{is}\in\{0,1\}, \\
& \sum_{o}x_{io}+\sum_{g}y_{ig}+\sum_{s}z_{is}=1, \\
& 0\le p_{i,t}\le \sum_s P_s^{\max}z_{is}.
\end{split}
\end{equation}
Here, $x$, $y$, $z$ are binary variables indicating the choice of serving, repositioning, or charging, respectively. The constraint $\sum x + \sum y + \sum z = 1$ ensures exactly one task is chosen per vehicle.

\subsection{Policy parameterization (intentions)}
Instead of outputting the final action $a_{i,t}$ directly (which requires satisfying complex global constraints like feeder limits), our neural policy samples \emph{intentions} $\tilde a_{i,t}$. The policy factorizes into discrete mode selection and continuous power control:
\begin{equation}
\label{eq:intention}
\begin{split}
\pi_\theta(a_{i,t}|s_t) ={}& \pi_\theta(m_{i,t}|s_t) \\
& \cdot \pi_\theta(\text{target}_{i,t}|m_{i,t},s_t) \\
& \cdot \pi_\theta(p_{i,t}|m_{i,t}=\text{chg},s_t),
\end{split}
\end{equation}
where $m_{i,t}\in\{\text{serve},\text{reb},\text{chg}\}$ is a high-level mode. These ``soft'' intentions are then passed to the MILP projection layer (Sec.~\ref{sec:milp}) to be converted into physically feasible ``hard'' actions.

\subsection{Reward}

We retain exactly four components to balance revenue, costs, and service quality:
\begin{equation}
\label{eq:reward}
\begin{aligned}
r_t(s_t,a_t) ={}&
\underbrace{\sum_{o\in \text{served}_t} R_t(h_o^{\text{pu}},h_o^{\text{do}})}_{\text{trip revenue}}
-
\underbrace{\sum_{i\in\mathcal V} c^{\text{drv}}\, d_{i,t}}_{\text{driving/ops cost}}
\\[2pt]
&-
\underbrace{\sum_{i\in\mathcal V} p^{\text{elec}}_{s(i),t}\,p_{i,t}\,\Delta t}_{\text{electricity}}
\\[2pt]
&-
\underbrace{\lambda_{\text{wait}}\sum_{o\in\mathcal O_t}\mathrm{wait}_o
\;+\;
\lambda_{\text{drop}}\bigl|\mathrm{dropped}_t\bigr|}_{\text{customer service.}}
\end{aligned}
\end{equation}

Here, $R_t(h_o^{\text{pu}},h_o^{\text{do}})$ is the revenue from trip $o$, $c^{\text{drv}}$ is the per-distance driving cost, $d_{i,t}$ is the distance traveled by vehicle $i$, $p^{\text{elec}}_{s(i),t}$ is the electricity price at station $s(i)$, $\mathrm{wait}_o$ is the waiting time for order $o$,
$\lvert\mathrm{dropped}_t\rvert$ is the number of dropped orders, and
$\lambda_{\text{wait}}$, $\lambda_{\text{drop}}$ are weighting coefficients.

\subsection{Semi-MDP dynamics and physics}
The system evolution is governed by the variable duration $\Delta(a_t)$ of the selected action:
\begin{equation}
\label{eq:durations}
\begin{split}
\Delta_{\text{serve}} &= T_t(h_{i,t},h_o^{\text{pu}})+T_t(h_o^{\text{pu}},h_o^{\text{do}}), \\
\Delta_{\text{reb}} &= T_t(h_{i,t},g), \\
\Delta_{\text{chg}} &= 1.
\end{split}
\end{equation}
Charging is modeled as a 1-step action ($\Delta_{\text{chg}}=1$) to allow frequent re-evaluation of power levels, whereas trips are atomic actions spanning multiple steps.

The physical state update equations are:
\begin{equation}
\label{eq:physics}
\begin{aligned}
&\text{Serve: }h_{i,t+\Delta_{\text{serve}}}=h_o^{\text{do}},\\
&\qquad E_{i,t+\Delta_{\text{serve}}}=E_{i,t}-\eta_{\text{drv}}\big(e^{\text{pk}}+e^{\text{trip}}\big),\\[2pt]
&\text{Reposition: }h_{i,t+\Delta_{\text{reb}}}=g,\\
&\qquad E_{i,t+\Delta_{\text{reb}}}=E_{i,t}-\eta_{\text{drv}}\,e^{\text{rb}}(h_{i,t},g),\\[2pt]
&\text{Charge: }h_{i,t+1}=h_{i,t},\\
&\qquad E_{i,t+1}=\min\{E_i^{\max},\ E_{i,t}+\eta_c\,p_{i,t}\Delta t\}.
\end{aligned}
\end{equation}
Here $e^{\text{rb}}(h_1,h_2)$ denotes the energy required for a reposition leg from $h_1$ to $h_2$. Longer charging sessions are represented by repeating the charge action across multiple consecutive steps.

Aggregated idle vehicle counts $n^{\text{idle}}_{h,t}$ evolve according to a hex-level flow conservation principle balancing departures and arrivals from serving, repositioning, and completed charging; in the simulator, these counts are updated by explicitly tracking vehicles.

\subsection{Duration-aware soft backup}
In a standard MDP, future rewards are discounted by $\gamma$. In our semi-MDP, the discount factor depends on the action duration $\Delta(a_t)$, leading to the non-robust soft target
\begin{equation}
\label{eq:backup_nonrobust}
\begin{split}
y_t&=r_t+\gamma^{\Delta(a_t)}\ \E_{a'\sim\pi}\Big[\min_k Q_{\bar\psi_k}(s_{t+\Delta(a_t)},a')\\
&\quad-\alpha\log\pi(a'|s_{t+\Delta(a_t)})\Big],
\end{split}
\end{equation}
where $(Q_{\bar\psi_k})$ are slowly updated target critics. This ensures that longer trips are discounted more heavily, correctly reflecting the time value of money and opportunity cost.

\section{Feasible-Action Projection via Rolling MILP}
\label{sec:milp}

\noindent Standard RL policies may output actions that violate complex coupled constraints (e.g., total power of all vehicles exceeding the grid feeder limit). To guarantee safety, we introduce a feasible-action projection layer $\Pi_{\text{feas}}$.
Given the intention vector $\tilde a_t$ from the policy, we compute the closest feasible action $a_t$ by solving a mixed-integer linear program (MILP).

\subsection{Time-limited MILP formulation}
Let $v$ stack all decision variables. The MILP objective seeks to maximize revenue while staying close to the policy's intention:
\begin{equation}
\label{eq:milp_obj}
\max_{v}\ \ \sum_{i,o}R_t(o)x_{io}
-\sum_i c^{\text{drv}} d_{i,t}
-\sum_{i,s} p^{\text{elec}}_{s,t}\,p_{i,s}\,\Delta t
-\mu\|v-\tilde a_t\|_1.
\end{equation}
This maximization is subject to critical operational constraints:

1.\ \textbf{Assignment logic:} Each vehicle performs exactly one task, and each order is served by at most one vehicle:
\begin{equation}
\label{eq:milp_assign}
\sum_o x_{io}+\sum_g y_{ig}+\sum_s z_{is}=1,\qquad \sum_i x_{io}\le 1.
\end{equation}

2.\ \textbf{Infrastructure limits:} The number of charging vehicles cannot exceed port capacity $C_s^{\text{port}}$, and the aggregate charging power must respect the feeder limit $P_{\text{feed}}^{\max}$:
\begin{equation}
\label{eq:milp_ports}
\sum_i z_{is}\le C_s^{\text{port}},\quad \sum_{i,s} p_{i,s}\le P_{\text{feed}}^{\max},\quad 0\le p_{i,s}\le P_s^{\max}z_{is}.
\end{equation}
The feeder constraint couples all vehicles across stations.

3.\ \textbf{Positive charging:} If a vehicle chooses to charge ($z_{is}=1$), it must draw a minimum non-zero power $p_{\min}$ to prevent ``fake'' charging actions:
\begin{equation}
\label{eq:pmin}
p_{i,s}\ \ge\ p_{\min}\,z_{is}.
\end{equation}

4.\ \textbf{Energy feasibility:} The SoC at the next step must remain within safety bounds $[E^{\min}, E^{\max}]$ after accounting for consumption or charging. Let $e^{\text{od}}_{io}$ denote the total distance for order $o$ (pickup plus trip) if served by vehicle $i$, and $e^{\text{rb}}_{ig}$ the distance for reposition from $h_{i,t}$ to $g$. We enforce
\begin{equation}
\label{eq:milp_soc}
\begin{split}
E_{i,t+1}&=E_{i,t}
          -\eta_{\text{drv}}\sum_o e^{\text{od}}_{io}\,x_{io}
          -\eta_{\text{drv}}\sum_g e^{\text{rb}}_{ig}\,y_{ig}\\
&\hspace{10mm}
          +\eta_c\sum_s p_{i,s}\Delta t,\\
E_i^{\min}\ &\le\ E_{i,t+1}\ \le\ E_i^{\max}.
\end{split}
\end{equation}
The projected total charging power is $p_i=\sum_s p_{i,s}$.

Solving this MILP at every step guarantees that the deployed action $a_t$ is physically valid. The projection operator is defined as
\[
a_t = \Pi_{\text{feas}}(s_t,\tilde a_t),
\]
where the MILP is solved using a branch-and-bound solver under a fixed time limit \(\tau_{max}\). During the search, the solver maintains an incumbent, defined as the best feasible solution found so far. If at least one feasible solution is identified before timeout, the current incumbent is returned; otherwise, a greedy fallback rule is applied to ensure that a feasible action is always produced.

\begin{algorithm}[tb]
\caption{Greedy Fallback Projection}
\label{alg:greedy_fallback}
\begin{algorithmic}[1]

\STATE Initialize empty action vector $a_t$

\STATE Initialize state $s_t$, intention $\tilde a_t$, residual feeder capacity $P_{\mathrm{rem}} \leftarrow P^{\max}_{\mathrm{feed}}$

\STATE Sort vehicles by increasing SoC $E_{i,t}$

\FOR{each vehicle $i$}
    \STATE Construct feasible action set $C_{i,t}$
    
    \IF{exists feasible serving order}
        \STATE Select order with highest immediate reward
    \ELSIF{exists feasible charging station with $P_{rem} \geq p_{min}$}
        \STATE Select nearest station $s$ with available port
        \STATE Allocate power $p_{i,s} \leftarrow \min(P^{\max}_s, P_{\mathrm{rem}})$
        \STATE $P_{\mathrm{rem}} \leftarrow P_{\mathrm{rem}} - p_{i,s}$
    \ELSE
        \STATE Select best energy-feasible reposition or idle action 
        \ENDIF 
    \ENDFOR
\RETURN assembled action vector $a_t$

\end{algorithmic}
\end{algorithm} 
\subsection{Greedy Fallback Projection}

In rare cases where the MILP solver fails to identify any feasible
solution within the time limit, no incumbent is available. To ensure
that a valid action is always produced, we activate a deterministic
greedy fallback procedure.

The fallback assigns decisions sequentially to vehicles in ascending order of state-of-charge (SoC), prioritizing energy-critical vehicles. For each vehicle, the algorithm selects the best feasible action subject to remaining infrastructure and energy constraints. We assume that a stay-idle action with zero energy consumption is always available in the feasible action set, which guarantees that at least one valid action exists for every vehicle at each decision step.

Specifically, the procedure operates as Algorithm \ref{alg:greedy_fallback}.

\subsection{Feasibility Guarantee}

By construction, the projection operator $\Pi_{\mathrm{feas}}$
returns either (i) a MILP incumbent satisfying constraints (13)--(16),
or (ii) a solution generated by the greedy fallback that explicitly
enforces these constraints. Therefore, $\Pi_{\mathrm{feas}}$ always
produces a physically admissible action.
\section{Graph-Aligned Wasserstein Robustness}
\label{sec:mag}

\noindent The transition dynamics of our semi-MDP depend on the exogenous
scenario $\xi = (D, T)$ introduced in Sec.~\ref{sec:system}. To prevent
the policy from overfitting to the empirical training distribution
$\hat P$ of such scenarios, we optimize its performance against a
worst-case distribution $P$ within a Wasserstein ambiguity set.

\subsection{Graph-aligned Mahalanobis ground metric}
Standard Wasserstein DRO uses the Euclidean distance $\|\xi - \xi'\|_2$
(e.g., \cite{Esfahani2018}), which treats all scenario dimensions equally.
In a city grid, perturbations should respect the spatial topology (e.g., demand shifting to a neighboring hex is more likely than shifting to a distant one).
We define a graph-aligned metric $d_Q$:
\begin{equation}
\label{eq:dQ}
d_Q(\xi,\xi')=\big\|Q^{1/2}(\xi-\xi')\big\|_2 = \norm{\xi-\xi'}_Q,
\end{equation}
where the precision matrix $Q$ is constructed as
\[
Q=\diag(w)+\beta\,Q_{\text{graph}}.
\]
Here, $\diag(w)$ scales features by their inverse variance, and $Q_{\text{graph}}$ is a Laplacian-based matrix encoding the hex grid and OD graph structure. This metric penalizes spatially disjoint perturbations more heavily than smooth, local variations.

\subsection{Wasserstein ambiguity set}
We define the ambiguity set $\mathcal P_\rho(\hat P)$ as the set of all distributions $P$ that are within a Wasserstein-1 distance $\rho$ from the empirical distribution $\hat P$:
\begin{equation}
\mathcal P_\rho(\hat P)=\{\,P\in\mathcal P(\R^q):\ W^{d_Q}_1(P,\hat P)\le \rho\,\}.
\end{equation}

\subsection{Robust backup via Kantorovich--Rubinstein dual}
Given a pre-decision state $s_t$ and projected action
$a_t = \Pi_{\mathrm{feas}}(s_t,\tilde a_t)$, we denote by
$s'(\xi)$ and $\Delta(\xi)$ the successor state and action duration
obtained by simulating the semi-MDP dynamics in Sec.~\ref{sec:semi_mdp}
under scenario $\xi$ (so $s'(\hat\xi_t)=s_{t+\Delta(a_t)}$ and
$\Delta(\hat\xi_t)=\Delta(a_t)$).
The robust value target at a transition evaluates the worst-case expected value:
\begin{equation}
\label{eq:backup_robust_KR_inf}
y_t^{\text{rob}}
= r_t + \inf_{P \in \mathcal P_\rho(\hat P)} \E_P\big[\gamma^{\Delta(\xi)}V_\varphi(s'(\xi))\big].
\end{equation}

By utilizing the identity $\inf(V) = -\sup(-V)$ and applying the Kantorovich--Rubinstein dual representation of the Wasserstein-1 ball to the cost $-V$ (see, e.g., \cite{Esfahani2018, Sinha2018}), we obtain the exact dual formulation:
\begin{equation}
\label{eq:backup_robust_KR}
\begin{aligned}
y_t^{\text{rob}}
&= r_t + \sup_{\lambda\ge 0}\Big\{-\lambda\rho + \E_{\hat P}\big[\inf_{\xi}\big(\gamma^{\Delta(\xi)}V_\varphi(s'(\xi)) \\
&\hspace{28mm}+ \lambda\, d_Q(\xi,\hat\xi)\big)\big]\Big\}.
\end{aligned}
\end{equation}

Our WDRO analysis assumes that the discounted value function is locally Lipschitz under $d_Q$. In our implementation, scenarios $\xi$ are restricted to a compact set $\mathcal{B}$, inputs are normalized, and $V_\varphi$ utilizes Lipschitz-continuous activations, ensuring this condition is practically satisfied.

In practice, we adopt a penalized, sample-based variant with a time-varying dual variable $\lambda_t$ and a bounded support set $\mathcal B$ around each empirical sample $\hat\xi_t$, yielding the practical robust target:
\begin{equation}
\label{eq:practical_target}
y_t^{\text{rob}}
= r_t
- \lambda_t\rho
+ \min_{\xi\in\mathcal B}
   \big[
      \gamma^{\Delta(\xi)} V_\varphi(s'(\xi))
      + \lambda_t\,d_Q(\xi,\hat\xi_t)
   \big].
\end{equation}

\subsection{Inner minimization via projected (sub)gradient descent}
To solve the inner minimization in \eqref{eq:practical_target}, we use projected (sub)gradient descent over $\xi$. Let
\[
f(\xi)=\gamma^{\Delta(\xi)}V_\varphi(s'(\xi)) + \lambda_t\,d_Q(\xi,\hat\xi_t).
\]
We treat $\Delta(\xi)$ as piecewise constant with respect to $\xi$ and ignore $\nabla_\xi \gamma^{\Delta(\xi)}$ in practice, leading to the update
\[
\xi^{k+1/2}
= \xi^k
- \eta\Big(
      \gamma^{\Delta(\xi^k)}\nabla_\xi V_\varphi(s'(\xi^k))
      + \lambda_t\,g_d(\xi^k)
    \Big),
\]
where $\nabla_\xi V_\varphi$ is obtained via backpropagation through the value network, and $g_d$ is the subgradient of the Wasserstein distance term:
\[
g_d(\xi)=
\begin{cases}
\dfrac{Q(\xi-\hat\xi_t)}{\max\{\varepsilon,\ \norm{\xi-\hat\xi_t}_{Q}\}}, & \xi\neq \hat\xi_t,\\[0.8ex]
\text{any }g:\ \norm{g}_{Q^{-1}}\le 1, & \xi=\hat\xi_t.
\end{cases}
\]
We then project $\xi^{k+1/2}$ back to the support set $\mathcal B$:
\[
\xi^{k+1} = \Proj_{\mathcal B}(\xi^{k+1/2}),
\]
e.g., via closed-form projection for $Q$-balls. Starting from $\xi^0 = \hat\xi_t$ and iterating $K$ steps yields an approximate maximizer $\xi^\star$ used in \eqref{eq:practical_target}.

\subsection{Primal--dual risk-budget tracking}
To control the \emph{effective} adversarial radius realized by the inner maximization, we track
\begin{equation}
\label{eq:rho_hat}
\hat\rho_t = d_Q(\xi_t^\star,\hat \xi_t),
\end{equation}
and update the dual variable $\lambda_t$ toward a prescribed risk budget $\rho_{\text{target}}\in(0,\rho]$:
\begin{equation}
\label{eq:lambda_update}
\lambda_{t+1}=\Big[\lambda_t+\eta_t\big(\hat\rho_t-\rho_{\text{target}}\big)\Big]_+,\qquad \eta_t=\frac{\eta_0}{\sqrt{t}},
\end{equation}
where $[\,\cdot\,]_+$ denotes projection onto $\R_+$. This primal--dual rule penalizes the adversary when the realized radius exceeds $\rho_{\text{target}}$ and relaxes it otherwise. Theorem~\ref{thm:PD1} (Sec.~\ref{sec:theory}) shows that the average budget violation vanishes as $t\to\infty$ under mild conditions.

\section{Learning Architecture and Training}
\label{sec:arch}

\begin{figure*}
    \centering \includegraphics[width=7.16in]{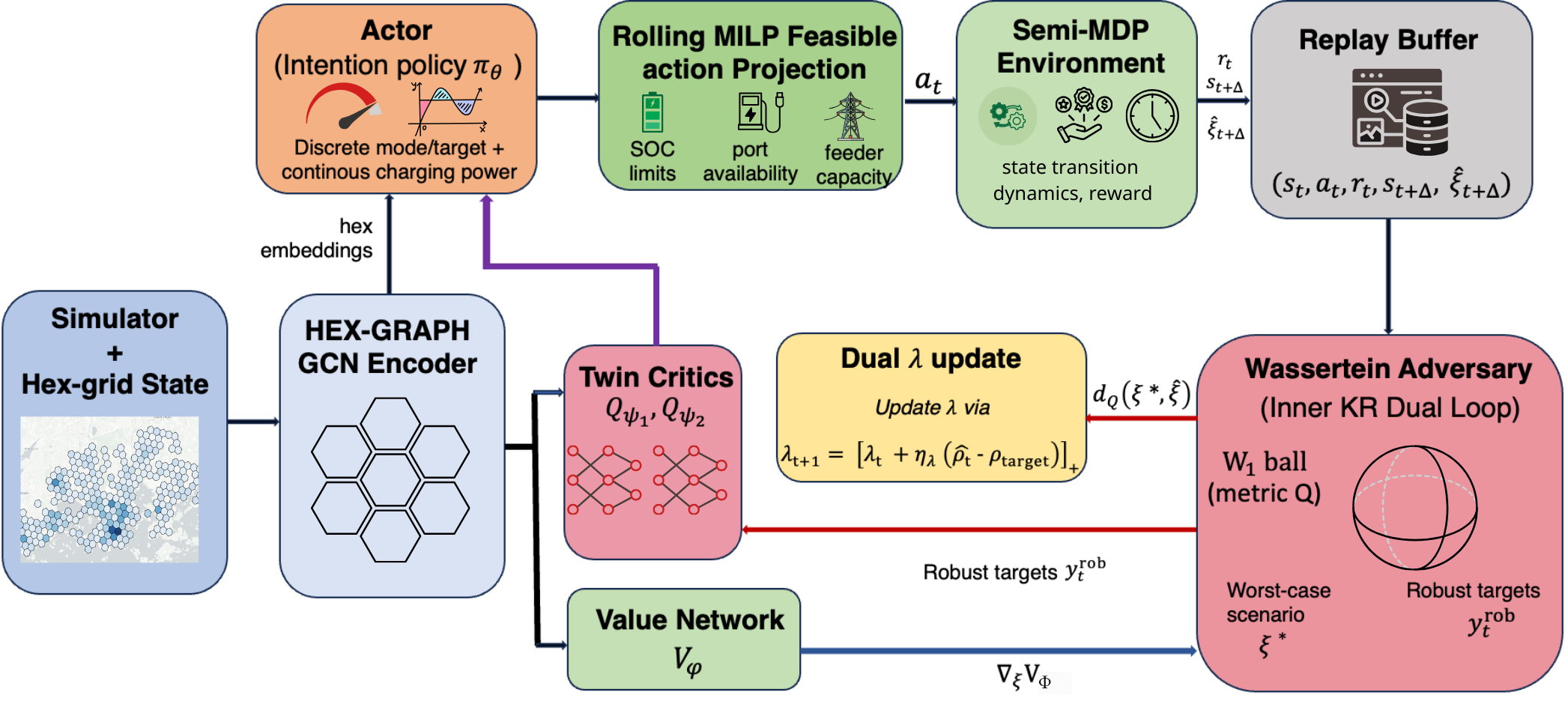}
    \caption{Overall PD--RSAC architecture. The simulator produces the aggregate hex-grid state, which is encoded by a shared GCN. The actor outputs mixed discrete--continuous intentions, which are projected by a rolling MILP into feasible actions before execution in the semi-MDP environment. Transitions are stored in a replay buffer. During training, the value network and Wasserstein adversary construct robust targets, while the twin critics and the primal--dual update refine the policy under distributional robustness.}
    \label{fig:arch}
\end{figure*}

\noindent At each decision epoch, the simulator produces the aggregated hex-grid
state $s_t$ and scenario $\hat\xi_t$, as illustrated in Fig.~\ref{fig:arch}.
A shared hex-graph GCN encoder maps per-hex features to embeddings that
are consumed by the actor, twin critics, and value network. The actor
samples per-vehicle intentions, and a rolling MILP projection layer
enforces SoC, port-capacity, and feeder constraints to obtain a feasible
system action $a_t = \Pi_{\mathrm{feas}}(s_t,\tilde a_t)$, which is
executed in the semi-MDP environment. The resulting transitions are stored
in a replay buffer and reused off-policy. For each sampled transition, the
value network and Wasserstein adversary run the inner KR-dual loop to
construct a worst-case scenario $\xi^\star$, yielding robust targets
$y_t^{\mathrm{rob}}$ that update the critics and the actor, while the dual
variable $\lambda$ is adapted to track the desired risk budget.

\subsection{Hex-graph encoder (GCN)}
We use a Graph Convolutional Network (GCN) \cite{Kipf2017} to extract spatial features. Let $A$ be the adjacency of the hex graph (with self-loops), $\tilde A = A + I$, $\tilde D = \diag(\sum_j \tilde A_{ij})$, and $\hat A = \tilde D^{-1/2} \tilde A\,\tilde D^{-1/2}$. The node features $\Phi$ (vehicle counts, demand, infra features, time encodings) are processed as
\begin{align}
H^{(1)} &= \sigma\!\big(\hat A\,\Phi\,W_0 + \mathbf{1}b_0^\top\big),\nonumber\\
E=H^{(2)} &= \sigma\!\big(\hat A\,H^{(1)}\,W_1 + \mathbf{1}b_1^\top\big), \label{eq:gcn}
\end{align}
yielding hex embeddings $E$ that inform both the actor and critics. In our implementation, $\sigma$ is the SiLU activation.

\subsection{Actor with Gumbel--Softmax and squashed Gaussian}
To allow backpropagation through the discrete action selection, we use the Gumbel--Softmax reparameterization \cite{Jang2017}. For a set of logits $(u_k)$, the discrete intention $\tilde z$ is sampled as:
\begin{equation}
\label{eq:gumbel}
\begin{split}
\tilde z_{k}&=\frac{\exp\big((u_{k}+g_k)/\tau\big)}{\sum_{j}\exp\big((u_{j}+g_{j})/\tau\big)},\\
g_k &\sim \mathrm{Gumbel}(0,1),
\end{split}
\end{equation}
where $\tau>0$ is a temperature annealed during training.

For the continuous charging power $\hat p$, we use a squashed Gaussian head conditioned on a charging context $h^{\text{chg}}_i$:
\begin{align}
(\mu_i,\log\sigma_i) &= \mathrm{MLP}(h^{\text{chg}}_i),\nonumber\\
u &\sim \mathcal N(\mu_i,\sigma_i^2),\qquad
\hat p=\frac{P_s^{\max}}{2}\big(1+\tanh u\big). \label{eq:power_head}
\end{align}
The log-density includes the standard change-of-variables correction due to the $\tanh$ squashing. The actor outputs an intention $\tilde a_t$ built from discrete Gumbel--Softmax samples and continuous $\hat p$.

\subsection{Critics and value network}
Twin critics $Q_{\psi_1},Q_{\psi_2}$ consume a state embedding (from the GCN) and an action embedding constructed from the \emph{projected} system action $a_t = \Pi_{\text{feas}}(s_t,\tilde a_t)$. A separate value network $V_\varphi$ shares the GCN encoder and outputs $V_\varphi(s'(\xi))$ used by the Wasserstein adversary in \eqref{eq:practical_target} as well as in the soft value definition below.

\subsection{Soft value and training losses}
Following the SAC formulation of \cite{Haarnoja2018},
the entropy-regularized value is defined as 
\begin{equation}
\label{eq:value_soft}
V_\varphi(s)=\E_{a\sim\pi_\theta(\cdot|s)}\big[\min_{k=1,2}Q_{\psi_k}(s,\Pi_{\text{feas}}(s,a))-\alpha\log \pi_\theta(a|s)\big].
\end{equation}
With the robust target $y_t^{\text{rob}}$ from \eqref{eq:practical_target}, we train the critics by minimizing
\begin{equation}
\label{eq:critic_loss}
\mathcal L_Q(\psi_k)=\E\big[(Q_{\psi_k}(s_t,a_t)-y_t^{\text{rob}})^2\big],\quad k=1,2,
\end{equation}
and the actor by the SAC-style  \cite{Haarnoja2018}:
\begin{equation}
\label{eq:actor_loss}
\mathcal L_\pi(\theta)=\E_{\!s,\tilde a\sim\pi_\theta}\Big[\alpha\log\pi_\theta(\tilde a|s)-\min_{\!k} Q_{\psi_k}\big(s,\Pi_{\text{feas}}(s,\tilde a)\big)\Big],
\end{equation}
where gradients through the projection $\Pi_{\text{feas}}$ are handled by a straight-through estimator (STE): in the forward pass we use $a=\Pi_{\text{feas}}(s,\tilde a)$, while in the backward pass we approximate $\partial a / \partial \tilde a \approx I$ along selected components. The composition of $\Pi_{\text{feas}}$ with the robust backup preserves the contraction property of the Bellman operator (Theorem~\ref{thm:T1}).

\subsection{Training algorithm: PD--RSAC}
The full training loop, named Primal--Dual Robust SAC (PD--RSAC), is summarized in Algorithm~\ref{alg:pdrsac}. Note that the primal--dual update \eqref{eq:lambda_update} is already defined in Sec.~\ref{sec:mag} and is simply invoked here.

\begin{algorithm}[tb]
\caption{PD--RSAC with Graph-Aligned $W_1$ and MILP Projection}
\label{alg:pdrsac}
\begin{algorithmic}[1]
\STATE Initialize actor $\theta$, critics $\psi_1,\psi_2$, value network $\varphi$, dual $\lambda\ge 0$, replay buffer $\mathcal D$.
\FOR{each environment step $t$}
    \STATE Observe $s_t$.
    \STATE Sample intention $\tilde a_t \sim \pi_\theta(\cdot|s_t)$ via \eqref{eq:intention}, \eqref{eq:gumbel}, \eqref{eq:power_head}.
    \STATE \textbf{Project:} $a_t \leftarrow \Pi_{\text{feas}}(s_t, \tilde a_t)$ via MILP \eqref{eq:milp_obj}--\eqref{eq:milp_soc}.
    \STATE Execute $a_t$ in the semi-MDP, observe $r_t, s_{t+\Delta}, \hat\xi_t$.
    \STATE Store $(s_t,\tilde a_t,a_t,r_t,s_{t+\Delta},\hat\xi_t)$ in $\mathcal D$.
    \STATE Sample minibatch of transitions from $\mathcal D$.
    \FOR{each transition in minibatch}
       \STATE Initialize $\xi^0 \leftarrow \hat\xi_t$.
       \FOR{$k=0$ to $K-1$}
          \STATE $g \leftarrow \gamma^{\Delta(\xi^k)} \nabla_\xi V_\varphi(s'(\xi^k))$.
          \STATE $h \leftarrow g_d(\xi^k)$ as in \eqref{eq:dQ}.
          \STATE $\xi^{k+1/2} \leftarrow \xi^k - \eta (g + \lambda_t h)$.
          \STATE $\xi^{k+1} \leftarrow \Proj_{\mathcal B}(\xi^{k+1/2})$.
       \ENDFOR
       \STATE $\xi^\star \leftarrow \xi^K$.
       \STATE Compute robust target $y_t^{\text{rob}}$ via \eqref{eq:practical_target}.
       \STATE Update critics by minimizing \eqref{eq:critic_loss}.
       \STATE Update actor by minimizing \eqref{eq:actor_loss}.
       \STATE Update $\lambda$ via the primal--dual rule \eqref{eq:lambda_update}.
    \ENDFOR
\ENDFOR
\end{algorithmic}
\end{algorithm}

\section{Theoretical Results}
\label{sec:theory}

\noindent We briefly summarize the theoretical properties underpinning PD--RSAC. 

\begin{assumption}[Boundedness \& discount]\label{ass:A1}
Rewards satisfy $\abs{r_t}\le R_{\max}<\infty$, and $0<\gamma<1$.
\end{assumption}

\begin{assumption}[Feasible projection]\label{ass:A2}
For every state $s$ and intention $\tilde a$, the projection operator $\Pi_{\mathrm{feas}}(s,\tilde a)$ returns a feasible action. This is guaranteed by combining the time-limited MILP \eqref{eq:milp_obj}--\eqref{eq:milp_soc} with a ``safe'' greedy fallback.
\end{assumption}

\begin{assumption}[Lipschitz regularity under $d_Q$]\label{ass:A3}
On the relevant domain, $V_\varphi(s'(\cdot))$ is $L_W$-Lipschitz under $d_Q(\cdot,\cdot)$.
\end{assumption}

\subsection{Contraction of the robust-soft Bellman operator}

For a fixed dual variable $\lambda\ge 0$, define the operator
\begin{equation}
\label{eq:bellman_rob}
\begin{aligned}
(\mathcal T_{\mathrm{rob}}V)(s)
= \E_{a\sim\pi(\cdot|s)}\Big[
   r(s,\hat a)
   - \alpha \log \pi(\hat a|s) \\
   \qquad\qquad
   + \sup_{\xi\in\mathcal B} \big(
       \gamma^{\Delta(\xi)} V(s'(\xi))
       - \lambda d_Q(\xi,\hat\xi)
     \big)
\Big].
\end{aligned}
\end{equation}
with $\hat a=\Pi_{\mathrm{feas}}(s,a)$ and $\Delta(\xi)\ge 1$ almost surely. The following property holds.

\begin{theorem}[Contraction of robust Bellman operator]
\label{thm:T1}
Under Assumption~\ref{ass:A1}, for any $V_1,V_2$,
\[
\norm{\mathcal T_{\mathrm{rob}}V_1-\mathcal T_{\mathrm{rob}}V_2}_\infty \le \gamma\,\norm{V_1-V_2}_\infty.
\]
Thus, $\mathcal T_{\mathrm{rob}}$ is a $\gamma$-contraction in the $L_\infty$ norm and admits a unique fixed point.
\end{theorem}

\begin{proof}
Rewards and entropy terms cancel in the difference $(\mathcal T_{\mathrm{rob}}V_1 - \mathcal T_{\mathrm{rob}}V_2)$. Using $\sup f - \sup g \le \sup (f-g)$ and the fact that $\gamma^{\Delta(\xi)} \le \gamma$ for all $\xi$ yields
\[
\abs{(\mathcal T_{\mathrm{rob}}V_1)(s)-(\mathcal T_{\mathrm{rob}}V_2)(s)}
\le \gamma\,\norm{V_1-V_2}_\infty
\]
for all $s$, proving the result.
\end{proof}

\subsection{\texorpdfstring{Lipschitz robustness bound for $W_1^{d_Q}$}{Lipschitz robustness bound for W1-dQ}}

We recall a standard Lipschitz robustness bound for Wasserstein balls
\cite{Esfahani2018}, specialized to $d_Q$.
\begin{theorem}[KR Lipschitz robustness bound]
\label{thm:T3}
Let $g:\R^q\to\R$ be $L_W$-Lipschitz under $d_Q$, i.e., $\abs{g(\xi)-g(\xi')} \le L_W d_Q(\xi,\xi')$. For the ambiguity set $\mathcal P_\rho(\hat P)$,
\[
\inf_{P\in\mathcal P_\rho(\hat P)} \E_P[g]
\ \ge\
\E_{\hat P}[g] - L_W \rho.
\]
In particular, the worst-case expected value over $\mathcal P_\rho(\hat P)$ is lower-bounded by the empirical expectation minus $L_W\rho$.
\end{theorem}

\begin{proof}
By Kantorovich--Rubinstein duality for $W_1^{d_Q}$,
\[
W_1^{d_Q}(P,\hat P)=\sup_{\norm{f}_{\mathrm{Lip}(d_Q)}\le1}(\E_P[f]-\E_{\hat P}[f]).
\]
Setting $f=g/L_W$ shows that for any $P$,
\[
\E_P[g]-\E_{\hat P}[g] \ge -L_W W_1^{d_Q}(P,\hat P) \ge -L_W \rho.
\]
Taking the infimum over $P\in\mathcal P_\rho(\hat P)$ yields the claim.
\end{proof}

Theorem~\ref{thm:T3} provides the theoretical motivation for the Wasserstein-robust design adopted in Section~\ref{sec:mag}. In particular, it shows that, whenever the value surrogate is Lipschitz under the graph-aligned metric \(d_Q\), the degradation of the worst-case expected value over the ambiguity set \(\mathcal P_\rho(\hat P)\) is controlled linearly by the radius \(\rho\). This justifies using \(\rho\) as an explicit robustness budget in the practical robust target \eqref{eq:practical_target} and in the primal--dual update \eqref{eq:lambda_update}.

\subsection{No-regret risk-budget tracking}

\begin{theorem}[No-regret tracking of the risk budget]
\label{thm:PD1}
Consider the primal--dual update
\[
\lambda_{t+1}=\big[\lambda_t+\eta_t g_t\big]_+,
\qquad
g_t=\hat\rho_t-\rho_{\text{target}},
\qquad
\eta_t=\frac{\eta_0}{\sqrt{t}},
\]
where \([\,\cdot\,]_+\) denotes projection onto \(\mathbb{R}_+\). Assume that \(\hat\rho_t\) is bounded on the compact support \(\mathcal{B}\), so that \(|g_t|\le G\) for some constant \(G>0\). Then the average budget violation satisfies
\[
\frac{1}{T}\sum_{t=1}^T (g_t)_+ = \mathcal O(T^{-1/2}),
\]
so that the average positive deviation of \(\hat\rho_t\) above \(\rho_{\text{target}}\) vanishes at rate \(\mathcal O(T^{-1/2})\).
\end{theorem}

\begin{proof}
The proof follows the standard projected subgradient argument for online convex optimization \cite{ShalevShwartz2012}, adapted to \eqref{eq:lambda_update}. Let \(\lambda^*\) be an optimal dual variable. By non-expansiveness of the projection \([\,\cdot\,]_+\),
\[
|\lambda_{t+1}-\lambda^*|^2
\le
|\lambda_t+\eta_t g_t-\lambda^*|^2.
\]
Expanding and rearranging gives
\[
2\eta_t g_t(\lambda^*-\lambda_t)
\le
|\lambda_t-\lambda^*|^2-|\lambda_{t+1}-\lambda^*|^2+\eta_t^2 g_t^2.
\]
Summing over \(t=1,\dots,T\) yields
\[
\sum_{t=1}^T g_t(\lambda^*-\lambda_t)
\le
\frac{|\lambda_1-\lambda^*|^2}{2\eta_T}
+\frac{1}{2}\sum_{t=1}^T \eta_t g_t^2.
\]
Since \(|g_t|\le G\), we have
\[
\sum_{t=1}^T g_t(\lambda^*-\lambda_t)
\le
\frac{|\lambda_1-\lambda^*|^2}{2\eta_T}
+\frac{G^2}{2}\sum_{t=1}^T \eta_t.
\]
With \(\eta_t=\eta_0/\sqrt{t}\), both \(\eta_T^{-1}\) and \(\sum_{t=1}^T \eta_t\) are \(\mathcal O(\sqrt{T})\), so the right-hand side is \(\mathcal O(\sqrt{T})\). Dividing by \(T\) gives an average violation bound of order \(\mathcal O(T^{-1/2})\), proving the claim.
\end{proof}
Intuitively, Theorem~3 guarantees that the primal--dual update on
$\lambda$ automatically calibrates the strength of the Wasserstein
penalty: when the adversary tends to exceed the desired radius
$\rho_{\text{target}}$, the dual variable grows and pushes the realized
radius back, and vice versa. As a result, the effective robustness level
remains close to $\rho_{\text{target}}$ over the course of training.

\section{Experiments}
\label{sec:experiments}

\subsection{Experimental Setup}
We evaluate PD--RSAC on a city-scale EV ride-hailing task built from the
New York City Taxi and Limousine Commission (TLC) Trip Record
Data~\cite{NYCTLC}. Trips are filtered for valid coordinates and
timestamps and restricted to central New York City (Manhattan and nearby
boroughs). The service region is tessellated into a hexagonal grid using
the H3 library~\cite{H3}; at resolution $8$ this yields $3824$ active hexes, each has the area of $0.73 \text{ km}^2$ and the radius of $0.46 \text{ km}$. There are 13,485,528 trips available, each trip is mapped to an origin and destination hex by its pickup and
drop-off coordinates. To match the fleet size of $N=1000$ vehicles, we sample
20\% of the daily trip data, leaving 2,045,463 trips for training and 651,643 trips eval for evaluation. For every $\Delta t = 5$ minutes, we aggregate
trips into an OD tensor $D_t(h_1,h_2)$ and estimate a discrete
travel-time field $T_t(h_1,h_2)$ from empirical median durations.

The hex adjacency used by the GCN encoder is obtained from H3 neighbor
relationships. Charging stations are placed at the $|\mathcal S|=150$ hexes using a demand-aware, spatially-spaced selection rule: hexes are ranked in descending order of historical pickup intensity and greedily selected subject to a 4-hop exclusion radius that enforces minimum spacing between stations, with any remaining slots filled by descending demand if the spacing constraint prevents reaching the target count, each with port capacity
$C_s^{\text{port}} =5$.
A global feeder limit $P_{\text{feed}}^{\max}= 7000$ is imposed on the total charging
power. Unless otherwise stated, the fleet consists of $V=1000$
homogeneous EVs with maximum capacity $E^{\max}=50$ kWh and initial SoCs drawn uniformly from
$[0.5,0.9]E^{\max}$. 
To evaluate the performance of our algorithm, the first 24 days of the dataset are used for training and the remaining 7 days are held out as the testing set, ensuring that the evaluation is conducted on unseen demand patterns. Each episode consists of 144 decision steps spanning 12. 

The reward function adopts net profit as the primary performance metric, defined as trip revenue minus driving and electricity costs, where driving cost is $\$0.30$ km and electricity cost is $\$0.18$ kWh; dropped-order and waiting-time penalties are weighted by $\lambda_{\text{drop}}=0.1$ and $\lambda_{\text{wait}}=0.5$, respectively.
The actor, twin critics, and value network are trained with the Adam optimizer using learning rates of $5\times10^{-5}$ (actor), $1\times10^{-4}$ (twin critics), and $1\times10^{-4}$ (value network), a discount factor of $\gamma=0.995$, a replay buffer capacity of $100{,}000$, and a mini-batch size of $128$. For the distributionally robust backup, the Wasserstein ambiguity radius is set to $\rho=0.3$ with a target risk budget of $\rho_{\text{target}}=0.2$, the graph-aligned precision matrix $Q$ uses $\beta=0.3$ to weight the Laplacian term, the primal--dual update on $\lambda$ is driven by an initial step size $\eta_0=0.01$, while the MILP projection layer is solved using Gurobi under a time limit of $\tau_{\max}=3.0$ seconds per decision epoch.  The
simulation experiments are conducted via Python 3.11 and Pytorch
source machine learning library. All simulation experiments are conducted on NVIDIA GeForce RTX 2080 Ti GPUs. For reproducibility, the full codebase and implementation details are available at \href{https://github.com/anthe8105/PD-RSAC.git}{https://github.com/anthe8105/PD-RSAC.git}.

\subsection{Compared Methods}
We evaluate four methods for EV fleet management under the same simulator, hex discretization, state representation, action space, network architecture, reward function, and training budget to ensure a fair comparison.

\begin{itemize}[leftmargin=*,itemsep=2pt,topsep=2pt,parsep=0pt,partopsep=0pt]
  \item \textbf{PD--RSAC (Proposed Method).}
  Our proposed method integrates Soft Actor-Critic with Wasserstein distributionally robust optimization and MILP-based feasible-action projection. This design combines learning-based policy optimization with optimization-based constraint enforcement to improve robustness under distributional uncertainty.
  
  \item \textbf{Greedy Heuristic.}
  A rule-based deterministic matching algorithm without learning. At each decision epoch, vehicles are assigned to trips by maximizing immediate profit under a maximum pickup distance constraint. Vehicles with state-of-charge below 20\% are prioritized for charging. An iterative greedy matching strategy is applied to maximize the number of served trips while satisfying energy and assignment constraints.

  \item \textbf{SAC (Soft Actor-Critic) \cite{Haarnoja2018}.}
  A standard off-policy actor--critic algorithm without the MILP-based feasible-action projection or Wasserstein distributionally robust optimization components.
  
  \item \textbf{MAPPO (Multi-Agent Proximal Policy Optimization) \cite{Yu2022}.}
  An on-policy multi-agent extension of PPO under the centralized training with decentralized execution paradigm. Each vehicle is modeled as an agent that learns a decentralized stochastic policy from local observations, while a centralized critic uses global fleet and grid information during training. The policy is optimized with PPO-style clipped surrogate objectives to improve training stability in cooperative multi-agent settings.

  \item \textbf{MADDPG (Multi-Agent Deep Deterministic Policy Gradient) \cite{Lowe2017}.}
  An off-policy multi-agent extension of DDPG under the centralized training with decentralized execution paradigm. Each vehicle is modeled as an agent that makes charging and dispatch decisions based on local observations. A deterministic actor produces decentralized actions, while a centralized critic takes joint observations and actions to estimate agent-specific value functions and mitigate non-stationarity during training. The policy is trained with replay-buffer-based updates and target networks to support stable learning in continuous control settings.
\end{itemize}

\subsection{Main Evaluation}

For PD--RSAC, the evaluation takes 10260 seconds for a week of simulation (2016 steps), which averages to approximately 5 seconds per 5-minute decision step. Although the per-step cost is dominated by the rolling MILP projection and the \(K\)-iteration inner loop of the Wasserstein adversary, this runtime remains well below the \(\Delta t = 300\)-second decision interval. This leaves ample headroom for real-time deployment and suggests that enforcing hard feasibility and distributional robustness does not compromise practical scalability at city scale.

\begin{figure}
\centering
\includegraphics[width=3.5in]{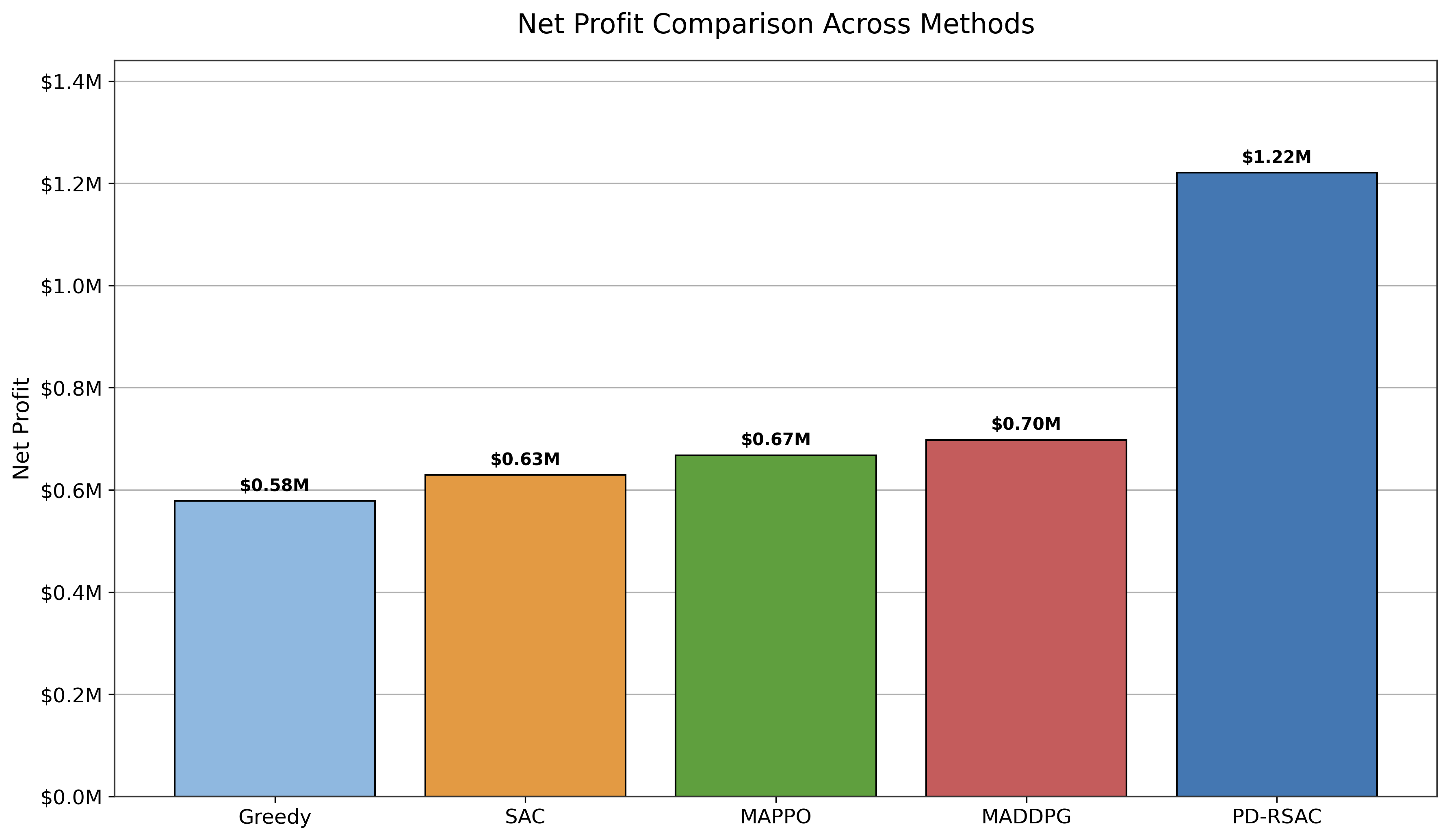}
    \caption{Net profit comparison across methods. PD--RSAC achieves the highest net profit among all baselines, substantially outperforming Greedy, SAC, MAPPO, and MADDPG.}
    \label{fig:figure1_net_profit}
\end{figure}

Figure~\ref{fig:figure1_net_profit} reports the net profit of all methods.
PD--RSAC achieves the strongest overall economic performance, reaching
\(\$1.22\text{M}\), substantially outperforming Greedy (\(\$0.58\text{M}\)),
SAC (\(\$0.63\text{M}\)), MAPPO (\(\$0.67\text{M}\)), and MADDPG (\(\$0.70\text{M}\)).
The gap is large across all baselines, indicating that the proposed method
consistently yields a more effective operational policy than both heuristic
and learning-based alternatives.

\begin{figure}
    \centering
    \includegraphics[width=3.5in]{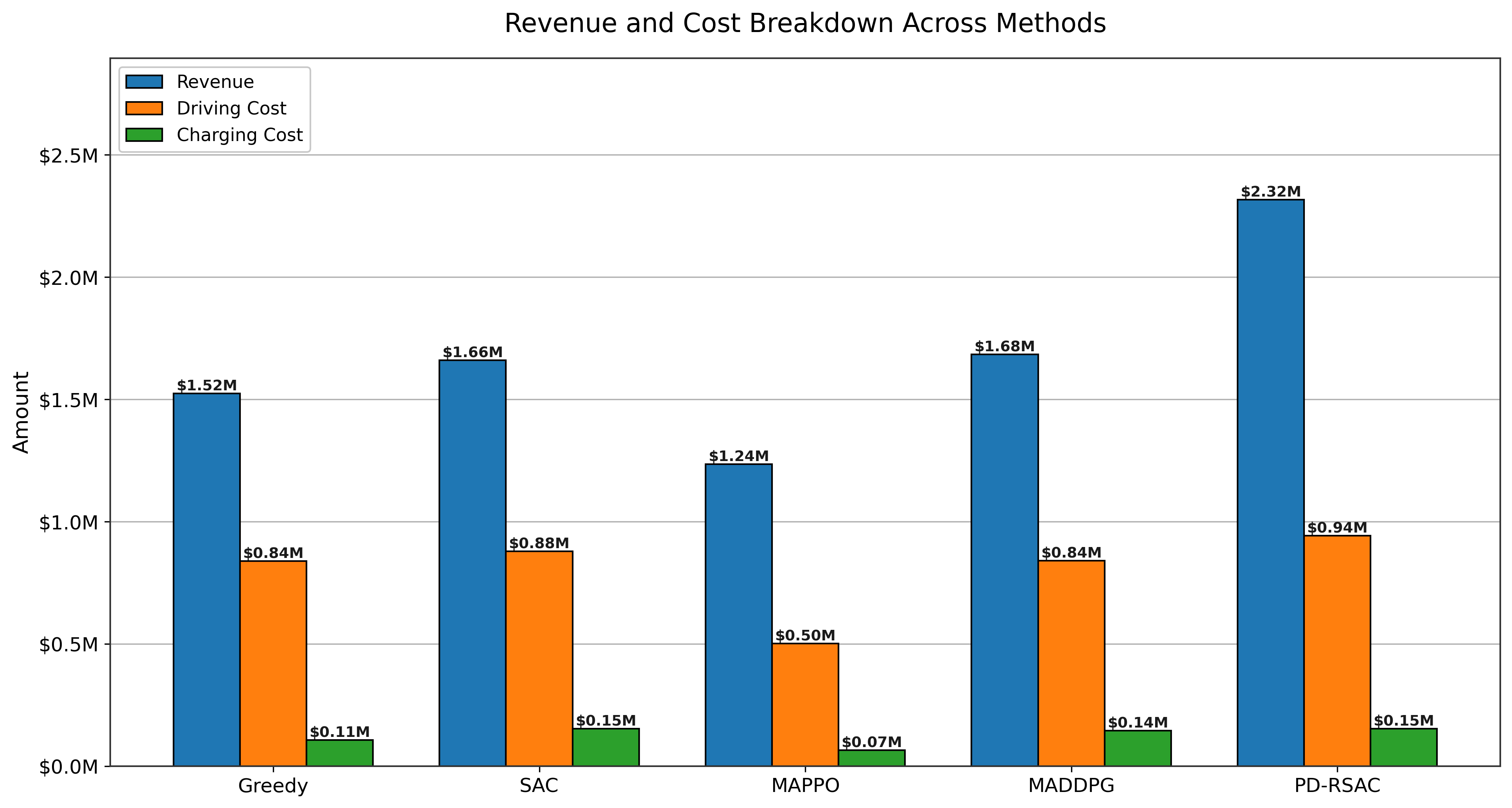}
    \caption{Revenue and cost breakdown across methods. PD--RSAC generates the highest total revenue, while its driving and charging costs remain within the same overall range as competing methods, yielding the best net profit.}
    \label{fig:figure2_financial_breakdown}
\end{figure}
To understand the source of this gain, Figure~\ref{fig:figure2_financial_breakdown}
shows the revenue and cost decomposition.
PD--RSAC attains the highest total revenue, \(\$2.32\text{M}\), again exceeding all baselines by a wide margin. Although its driving cost and charging cost
are not the lowest, the additional cost remains modest relative to the increase
in revenue. This is an important observation: PD--RSAC does not win merely by
suppressing cost, but by generating substantially more revenue through better
dispatching and charging decisions. In other words, the proposed method converts
available fleet and charging resources into profitable service more effectively
than the baselines.

\begin{figure}
    \centering
    \includegraphics[width=3.5in]{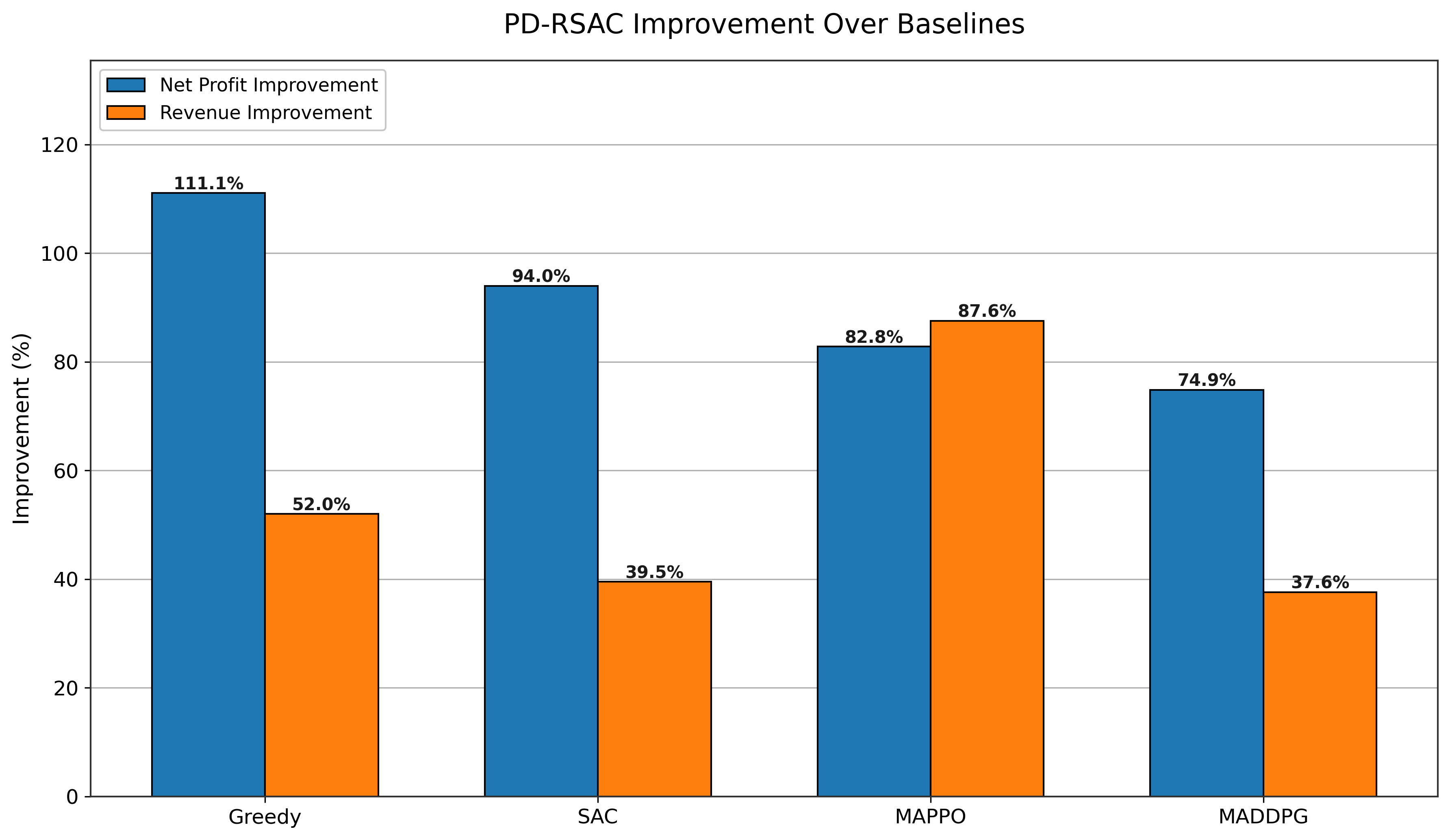}
    \caption{Relative improvement of PD--RSAC over baselines. The proposed method consistently improves both net profit and revenue across all compared baselines.}
    \label{fig:figure3_pdrsac_improvement}
\end{figure}
Figure~\ref{fig:figure3_pdrsac_improvement} further summarizes the relative
improvement of PD--RSAC over each baseline.
In terms of net profit, PD--RSAC improves over Greedy, SAC, MAPPO, and MADDPG
by \(111.1\%\), \(94.0\%\), \(82.8\%\), and \(74.9\%\), respectively.
The corresponding revenue improvements are \(52.0\%\), \(39.5\%\), \(87.6\%\),
and \(37.6\%\).
These gains are not isolated to one particular comparison, but remain consistent
across heuristic, single-agent RL, and multi-agent RL baselines.
Taken together, Figures~\ref{fig:figure1_net_profit}--\ref{fig:figure3_pdrsac_improvement}
show that PD--RSAC delivers the best overall economic outcome and that its advantage
is driven primarily by stronger revenue generation rather than by aggressive
cost reduction alone.

\begin{figure*}
    \centering
    \includegraphics[width=7.16in]{grid_safety_two_panel.png}
    \caption{Grid safety analysis under different controllers. (a) Compared with SAC, PD--RSAC eliminates feeder-limit violations by clipping charging peaks below the 7 MW constraint. (b) Compared with MAPPO, PD--RSAC remains safe while utilizing substantially more of the available feeder capacity, demonstrating a better safety--efficiency trade-off.}
    \label{fig:grid_safety_two_panel}
\end{figure*}
Beyond aggregate profit, we also examine whether this improved performance is obtained
in a safe and operationally meaningful manner.
Figure~\ref{fig:grid_safety_two_panel} compares grid power load profiles under
different controllers.

In Figure~\ref{fig:grid_safety_two_panel}(a), SAC produces substantial feeder-limit
violations, with a peak charging demand of approximately \(14{,}465\)~kW under a
\(7{,}000\)~kW feeder limit.
By contrast, PD--RSAC keeps the charging load below the limit throughout the day,
with a peak of \(6{,}999\)~kW.
This demonstrates that the proposed method can enforce the grid safety constraint
effectively even under high-demand operating conditions.

Figure~\ref{fig:grid_safety_two_panel}(b) provides a complementary comparison
against MAPPO.
Although MAPPO also remains within the feeder limit, its peak charging demand is
only about \(3{,}000\)~kW, corresponding to \(42.9\%\) of the available capacity.
PD--RSAC, in contrast, safely utilizes nearly the full feeder capacity while still
avoiding violations.
Hence, PD--RSAC is not only safe, but also less conservative and more efficient
in exploiting available grid resources.
This safety--efficiency trade-off is a central advantage of the proposed method:
it avoids the unsafe behavior observed in SAC while also avoiding the excessive
under-utilization exhibited by MAPPO.

\begin{figure}
    \centering
    \includegraphics[width=3.5in]{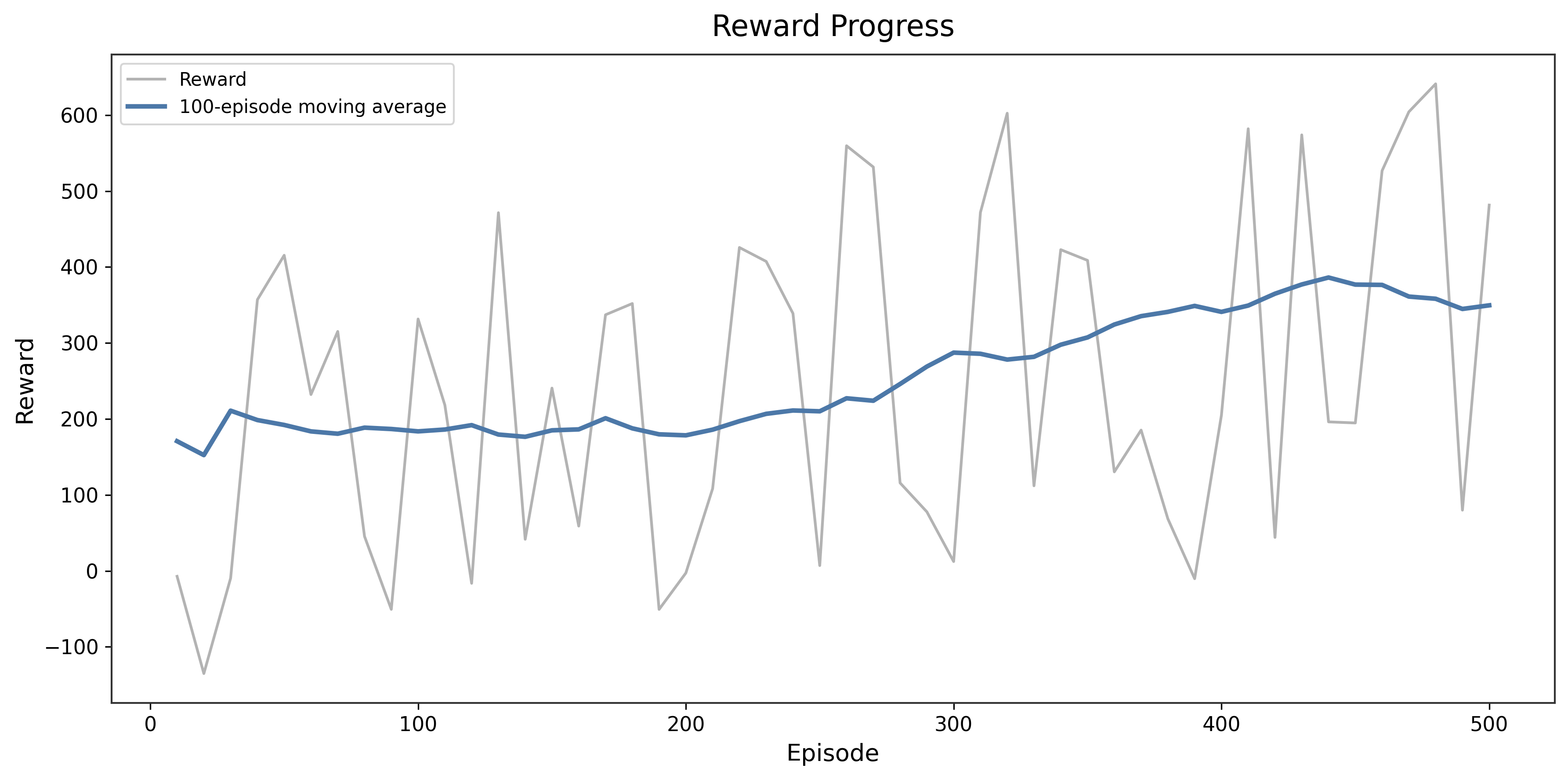}
    \caption{Training reward progression of PD--RSAC. Although the per-episode reward is noisy, the 100-episode moving average exhibits a clear upward trend, indicating stable policy improvement over time.}
    \label{fig:training_figure1_reward_progress}
\end{figure}
For completeness, we also inspect the training dynamics of PD--RSAC.
Figure~\ref{fig:training_figure1_reward_progress} shows that, although the per-episode
reward is noisy, the 100-episode moving average exhibits a clear upward trend over
training, indicating stable policy improvement.

\begin{figure}
\centering
\includegraphics[width=3.5in]{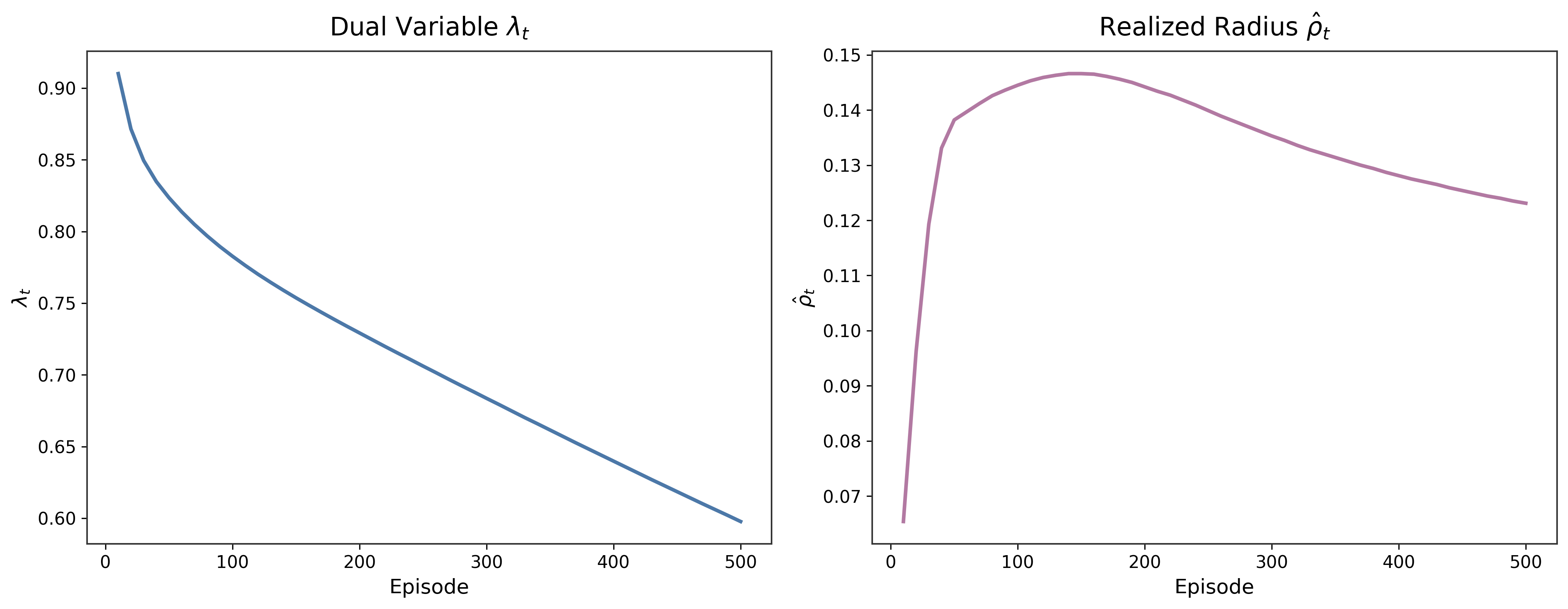}
    \caption{Evolution of the WDRO-specific variables during training. The dual variable \(\lambda_t\) decreases smoothly, while the realized radius \(\hat{\rho}_t\) increases in the early phase and then stabilizes, indicating a stable and meaningful robust training process.}
    \label{fig:training_figure3_wdro_mechanism}
\end{figure}
In addition, Figure~\ref{fig:training_figure3_wdro_mechanism} plots the evolution
of the dual variable \(\lambda_t\) and the realized radius \(\hat{\rho}_t\).
The dual variable decreases smoothly, while the realized radius increases in the
early phase and then gradually stabilizes.
This behavior suggests that the WDRO mechanism remains active and well-behaved during
training, adaptively calibrating the uncertainty set rather than collapsing to a
trivial solution or oscillating uncontrollably.

Overall, the main evaluation results support a consistent conclusion:
PD--RSAC achieves the highest economic performance among all evaluated methods,
learns stably during training, and enforces feeder-level safety constraints without
becoming overly conservative.
The method therefore offers a strong combination of profitability, robustness,
and operational safety.

\subsection{Ablation Study}

To isolate the contributions of individual components, we conduct an ablation study
by progressively removing the MILP projection layer, the Wasserstein robustness module,
and the graph-aligned ground metric.

Specifically, we consider the following variants:

\begin{itemize}[nosep]
    \item \textbf{w/o Graph}: replaces the graph-aligned \(Q\)-metric with the identity matrix, resulting in an unstructured Wasserstein ground cost.
    \item \textbf{w/o WDRO}: disables the Wasserstein robustness module and trains the agent using standard SAC without distributional robustness.
    \item \textbf{w/o MILP}: removes the MILP-based feasibility projection while retaining the robustness module and the graph-aligned \(Q\)-metric.
    \item \textbf{Full}: the complete PD-RSAC framework with MILP projection, Wasserstein DRO, and graph-aligned \(Q\)-metric.
\end{itemize}
For fair comparison, all variants are trained under identical hyperparameter settings and share the same ambiguity set radius.

\begin{figure*}
\centering
\includegraphics[width=7.16in]{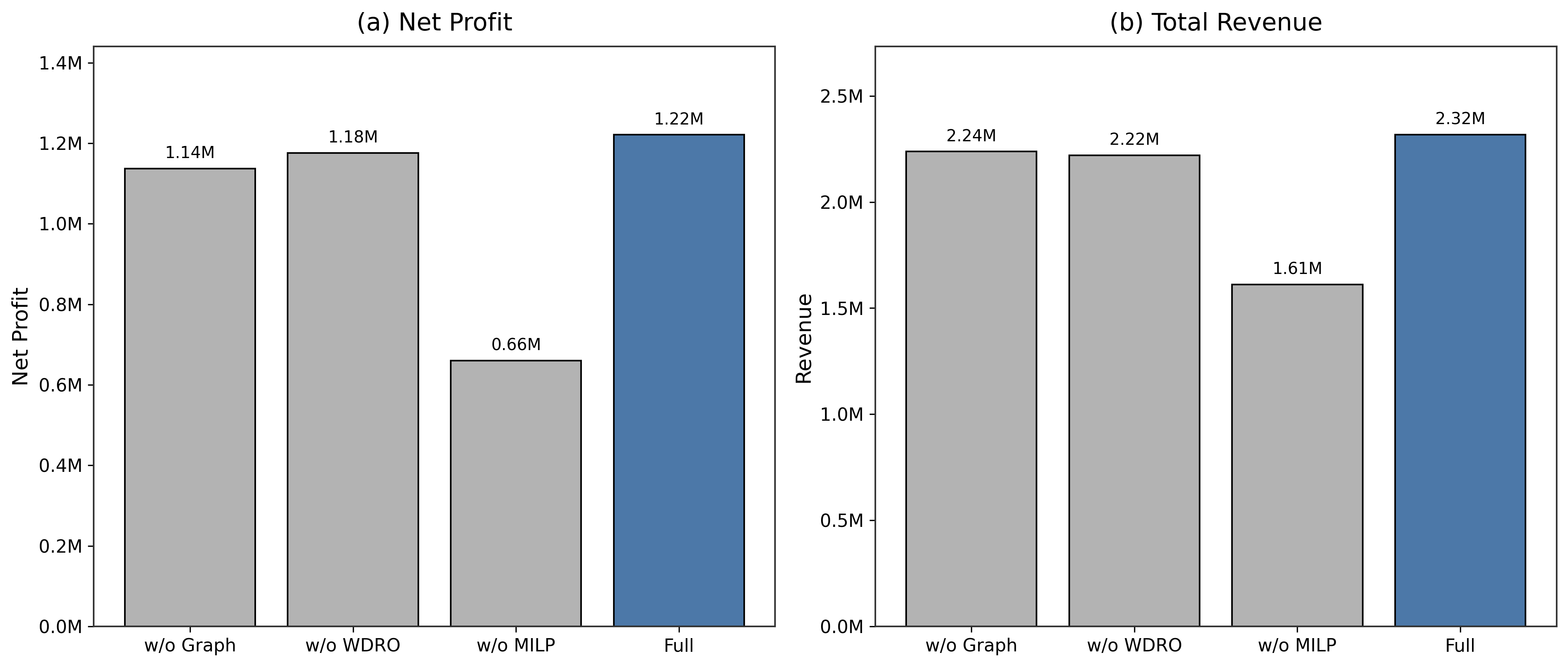}
\caption{Ablation results for PD--RSAC. The full model achieves the best performance in both net profit and total revenue. Removing MILP causes the largest degradation, while removing WDRO or the graph-aligned metric also leads to consistent performance drops.}
\label{fig:ablation_figure_main}
\end{figure*}

Figure~\ref{fig:ablation_figure_main} summarizes the ablation results in terms of
net profit and total revenue.
The full model achieves the best performance on both metrics, confirming that the
final gains of PD--RSAC do not come from a single module alone.
Rather, feasibility refinement, distributional robustness, and graph-aware geometry
each contribute to the final result.

Among all ablations, removing the MILP layer causes the largest degradation.
Without MILP projection, net profit drops from \(\$1.22\text{M}\) to \(\$0.66\text{M}\),
and total revenue decreases from \(\$2.32\text{M}\) to \(\$1.61\text{M}\).
This sharp decline indicates that the MILP module is a core component of the system.
Its role is not merely post-processing or safety filtering; instead, it materially
improves the realizability and quality of fleet decisions under charging and
infrastructure constraints.
These results suggest that feasibility-aware refinement is essential for translating
policy outputs into high-quality operational actions.

Removing the graph component also reduces performance.
The \textit{w/o Graph} variant reaches \(\$1.14\text{M}\) net profit and
\(\$2.24\text{M}\) revenue, both below the full model.
This shows that incorporating relational and spatial structure into the robustness
mechanism is beneficial.
Using an unstructured identity cost weakens the quality of the transportation-aware
uncertainty model, whereas the graph-aligned metric provides more meaningful
perturbation geometry for the decision problem.

Similarly, disabling WDRO leads to a measurable performance drop.
The \textit{w/o DRO} variant attains \(\$1.18\text{M}\) net profit and
\(\$2.22\text{M}\) revenue, again below the full model.
Although this degradation is smaller than that caused by removing MILP, it is still
consistent and non-trivial.
This indicates that Wasserstein distributional robustness provides additional gains
beyond nominal learning, improving the economic quality of the learned policy under
uncertainty.

Taken together, the ablation results confirm that all three components contribute
positively, with the MILP module having the largest impact and the graph-aware
representation and WDRO regularization providing complementary improvements.
Therefore, the strong performance of PD--RSAC is best understood as the result of
a tightly integrated design rather than a single dominant heuristic or architectural choice.

\begin{table}[!ht]
\centering
\caption{Training time comparison (seconds) across model variants.}
\label{tab:computing_time}
\begin{tabular}{lcc}
\hline
\textbf{Variant} & \textbf{Training time (s)} \\
\hline
w/o Graph & 516154.4 \\
w/o WDRO  & 496485.0 \\
w/o MILP  & 87898.1 \\
Full      & 600499.2 \\
\hline
\end{tabular}
\end{table}

Beyond effectiveness, it is also important to understand the computational overhead
introduced by each component. To this end, Table~\ref{tab:computing_time} reports the
training time of all ablation variants, complementing the performance comparison in
Fig.~\ref{fig:ablation_figure_main}.

Table~\ref{tab:computing_time} and Fig.~8 together show a clear efficiency--performance trade-off. Among all variants, \textit{w/o MILP} is the fastest to train, requiring only 87898.1\,s versus 600499.2\,s for the full model, i.e., about 6.83$\times$ less training time. This indicates that the MILP projection layer is the primary source of computation cost of the framework. However, the same variant also suffers the largest drop in net profit and total revenue, showing that the extra computational cost of MILP is closely tied to its strong contribution to feasibility refinement and overall decision quality.

\section{Conclusion}

We studied city-scale EV ride-hailing control as a constrained sequential decision-making problem in which dispatching, repositioning, and charging decisions must remain feasible under battery, station, and grid limitations. To address this challenge, we proposed PD--RSAC, a unified framework that combines semi-Markov reinforcement learning, rolling MILP-based action projection, and Wasserstein distributional robustness on a graph-structured transportation domain. In contrast to nominal learning pipelines, the proposed method integrates policy learning, feasibility refinement, and robustness within a single control loop.

Our results show that this integration is practically important. PD--RSAC achieves the highest net profit (\$1.22M) among all evaluated baselines while maintaining safe operation under feeder-capacity constraints. The ablation study further clarifies the role of each component: the MILP layer provides the largest contribution to profit by turning raw policy outputs into executable infrastructure-aware decisions, while the Wasserstein robustness module and graph-aligned metric provide complementary gains under uncertainty. Taken together, these findings suggest that effective EV fleet control requires not only strong learning capability, but also explicit treatment of physical feasibility and uncertainty within the decision loop itself.

More broadly, the paper highlights a useful design principle for large-scale cyber-physical decision systems: when actions are tightly constrained by infrastructure, learning alone is often insufficient, and performance improves substantially when optimization-based feasibility correction and robustness mechanisms are embedded directly into the control architecture. In this sense, PD--RSAC provides a concrete example of how reinforcement learning and mathematical programming can be combined to achieve both safety and profitability at scale.

An important direction for future work is to replace the current rolling MILP projection with a lighter feasibility module that achieves comparable decision quality at a much lower computational cost. Developing such a mechanism would make the framework more scalable without sacrificing the main economic benefits brought by feasibility-aware refinement.

\begingroup
\sloppy
\hbadness=10000
\bibliographystyle{IEEEtran}
\bibliography{references}

@article{Wei2021,
  author    = {Wei, K. and Vaze, V. and Jacquillat, A.},
  title     = {{Transit planning optimization under ride-hailing competition and traffic congestion}},
  journal   = {Transportation Science},
  volume    = {56},
  number    = {3},
  pages     = {725--749},
  year      = {2021},
  doi       = {10.1287/trsc.2021.1068}
}

@article{Cao2021,
  author    = {Cao, Y. and Wang, S. and Li, J.},
  title     = {The optimization model of ride-sharing route for ride hailing considering both system optimization and user fairness},
  journal   = {Sustainability},
  volume    = {13},
  number    = {5},
  pages     = {2728},
  year      = {2021},
  doi       = {10.3390/su13020902}
}

@article{Miao2015,
  author    = {Miao, F. and Han, S. and Lin, S. and Stankovic, J. A. and Zhang, D. and Munir, S. and Huang, H. and He, T. and Pappas, G. J.},
  title     = {Taxi dispatch with real-time sensing data in metropolitan areas: A receding horizon control approach},
  journal   = {IEEE Transactions on Automation Science and Engineering},
  volume    = {13},
  number    = {2},
  pages     = {463--478},
  year      = {2015},
  doi       = {10.1145/2735960.2735961}
}

@article{Liu2022,
  author    = {Liu, Y. and Wu, F. and Lyu, C. and Li, S. and Ye, J. and Qu, X.},
  title     = {Deep dispatching: A deep reinforcement learning approach for vehicle dispatching on online ride-hailing platform},
  journal   = {Transportation Research Part E: Logistics and Transportation Review},
  volume    = {161},
  pages     = {102694},
  year      = {2022},
  doi       = {10.1016/j.tre.2022.102694}
}

@article{Qin2020,
  author    = {Qin, Z. and Tang, X. and Jiao, Y. and Zhang, F. and Xu, Z. and Zhu, H. and Ye, J.},
  title     = {Ride-hailing order dispatching at DiDi via reinforcement learning},
  journal   = {INFORMS Journal on Applied Analytics},
  volume    = {50},
  number    = {5},
  pages     = {272--286},
  year      = {2020},
  doi       = {10.1287/inte.2020.1047}
}

@inproceedings{Yue2024,
  author    = {Yue, X. and Liu, Y. and Shi, F. and Luo, S. and Zhong, C. and Lu, M. and Xu, Z.},
  title     = {An end-to-end reinforcement learning based approach for micro-view order-dispatching in ride-hailing},
  booktitle = {Proceedings of the 33rd ACM International Conference on Information and Knowledge Management},
  pages     = {321--330},
  year      = {2024},
  publisher = {ACM},
  doi       = {10.1145/3627673.3680013}
}

@inproceedings{Wang2025,
  author    = {Jingwei Wang and Qianyue Hao and Wenzhen Huang and Xiaochen Fan and Qin Zhang and Zhentao Tang and Bin Wang and Jianye Hao and Yong Li},
  title     = {CoopRide: Cooperate All Grids in City-Scale Ride-Hailing Dispatching with Multi-Agent Reinforcement Learning},
  booktitle = {Proceedings of the 31st ACM SIGKDD Conference on Knowledge Discovery and Data Mining},
  pages     = {1--10},
  year      = {2025},
  doi       = {10.1145/3690624.3709205}
}

@article{Jiao2021,
  author    = {Yan Jiao and Xiaocheng Tang and Zhiwei Qin and Shuaiji Li and Fan Zhang and Hongtu Zhu and Jieping Ye},
  title     = {Real-world Ride-hailing Vehicle Repositioning using Deep Reinforcement Learning},
  journal   = {Transportation Research Part C: Emerging Technologies},
  volume    = {130},
  pages     = {103289},
  year      = {2021},
  doi       = {10.1016/j.trc.2021.103289}
}

@inproceedings{Li2022,
  author    = {Li, J. and Allan, V.},
  title     = {Where to go: Agent guidance with deep reinforcement learning in a city-scale online ride-hailing service},
  booktitle = {Proceedings of the IEEE International Conference on Intelligent Transportation Systems (ITSC)},
  pages     = {182--189},
  year      = {2022},
  doi       = {10.1109/ITSC55140.2022.9921747}
}

@article{Huang2022,
  author    = {Jianbin Huang and Longji Huang and Meijuan Liu and He Li and Qitian Tan and Xuesong Ma and Junjie Cui and De-Shuang Huang},
  title     = {Deep Reinforcement Learning-based Trajectory Pricing on Ride-hailing Platforms},
  journal   = {ACM Transactions on Intelligent Systems and Technology},
  volume    = {13},
  number    = {3},
  pages     = {1--19},
  year      = {2022},
  doi       = {10.1145/3474841}
}

@article{Jin2025,
  author    = {Jin, K. and Feng, Z. and Li, X. and Zhang, F.},
  title     = {Ride-hailing service pattern recognition and demand prediction},
  journal   = {IEEE Transactions on Intelligent Transportation Systems},
  volume    = {26},
  pages     = {1--10},
  year      = {2025},
  doi       = {10.1109/TITS.2025.3558274}
}

@article{Tian2025,
  author    = {Tian, Jingjing and Jia, Hongfei and Wang, Guanfeng and Huang, Qiuyang and Wu, Ruiyi and Gao, Heyao and Liu, Chao},
  title     = {Optimal scheduling of shared autonomous electric vehicles with multi-agent reinforcement learning: A MAPPO-based approach},
  journal   = {Neurocomputing},
  volume    = {622},
  pages     = {129343},
  year      = {2025},
  doi       = {10.1016/j.neucom.2025.129343}
}

@article{Hu2024,
  author    = {Hu, Jueming and Xu, Zhe and Wang, Weichang and Qu, Guannan and Pang, Yutian and Liu, Yongming},
  title     = {Decentralized graph-based multi-agent reinforcement learning using reward machines},
  journal   = {Neurocomputing},
  volume    = {564},
  pages     = {126974},
  year      = {2024},
  doi       = {10.1016/j.neucom.2023.126974}
}

@article{Kordabad2022,
  author    = {Kordabad, A. B. and Wisniewski, R. and Gros, S.},
  title     = {Safe Reinforcement Learning Using Wasserstein Distributionally Robust MPC and Chance Constraint},
  journal   = {IEEE Access},
  volume    = {10},
  pages     = {1--10},
  year      = {2022},
  doi       = {10.1109/ACCESS.2022.3228922}
}

@inproceedings{Achiam2017,
  author    = {Achiam, J. and Held, D. and Tamar, A. and Abbeel, P.},
  title     = {Constrained policy optimization},
  booktitle = {Proceedings of the 34th International Conference on Machine Learning (ICML)},
  pages     = {22--31},
  year      = {2017},
  doi       = {10.48550/arXiv.1705.10528}
}

@inproceedings{Amos2017,
  author    = {Amos, B. and Kolter, J. Z.},
  title     = {OptNet: Differentiable optimization as a layer in neural networks},
  booktitle = {Proceedings of the 34th International Conference on Machine Learning (ICML)},
  pages     = {152--161},
  year      = {2017},
  doi       = {10.48550/arXiv.1703.00443}
}

@inproceedings{Agrawal2019,
  author    = {Agrawal, A. and Barratt, S. and Boyd, S. and Busseti, E.},
  title     = {Differentiable convex optimization layers},
  booktitle = {Advances in Neural Information Processing Systems (NeurIPS)},
  volume    = {32},
  pages     = {1--10},
  year      = {2019},
  doi       = {10.48550/arXiv.1910.12430}
}

@article{Liu2023,
  author    = {Liu, Shihan and Yang, Xinyu and Zhang, Zhen and Lewis, Frank L.},
  title     = {Safe reinforcement learning for affine nonlinear systems with state constraints and input saturation using control barrier functions},
  journal   = {Neurocomputing},
  volume    = {518},
  pages     = {562--576},
  year      = {2023},
  doi       = {10.1016/j.neucom.2022.11.006}
}

@article{Iyengar2005,
  author    = {Iyengar, G. N.},
  title     = {Robust dynamic programming},
  journal   = {Mathematics of Operations Research},
  volume    = {30},
  number    = {2},
  pages     = {257--280},
  year      = {2005},
  doi       = {10.1287/moor.1040.0129}
}

@article{Nilim2005,
  author    = {Nilim, A. and El Ghaoui, L.},
  title     = {Robust control of Markov decision processes with uncertain transition matrices},
  journal   = {Operations Research},
  volume    = {53},
  number    = {5},
  pages     = {780--798},
  year      = {2005},
  doi       = {10.1287/opre.1050.0216}
}

@article{Esfahani2018,
  author    = {Mohajerin Esfahani, P. and Kuhn, D.},
  title     = {Data-driven distributionally robust optimization using the Wasserstein metric},
  journal   = {Mathematical Programming},
  volume    = {171},
  number    = {1},
  pages     = {115--166},
  year      = {2018},
  doi       = {10.1007/s10107-017-1172-1}
}

@inproceedings{Sinha2018,
  author    = {Sinha, A. and Namkoong, H. and Duchi, J.},
  title     = {Certifying some distributional robustness with principled adversarial training},
  booktitle = {Proceedings of the International Conference on Learning Representations (ICLR)},
  pages     = {1--10},
  year      = {2018},
  doi       = {10.48550/arXiv.1710.10571}
}

@inproceedings{Haarnoja2018,
  author    = {Haarnoja, T. and Zhou, A. and Abbeel, P. and Levine, S.},
  title     = {Soft actor-critic: Off-policy maximum entropy deep reinforcement learning with a stochastic actor},
  booktitle = {Proceedings of the 35th International Conference on Machine Learning (ICML)},
  pages     = {1861--1870},
  year      = {2018},
  doi       = {10.48550/arXiv.1801.01290}
}

@inproceedings{Jang2017,
  author    = {Jang, E. and Gu, S. and Poole, B.},
  title     = {Categorical reparameterization with Gumbel-Softmax},
  booktitle = {Proceedings of the International Conference on Learning Representations (ICLR)},
  pages     = {1--10},
  year      = {2017},
  doi       = {10.48550/arXiv.1611.01144}
}

@inproceedings{Kipf2017,
  author    = {Kipf, T. N. and Welling, M.},
  title     = {Semi-supervised classification with graph convolutional networks},
  booktitle = {Proceedings of the International Conference on Learning Representations (ICLR)},
  pages     = {1--10},
  year      = {2017},
  doi       = {10.48550/arXiv.1609.02907}
}

@inproceedings{GrandClement2021,
  author    = {Grand-Cl{\'e}ment, J. and Kroer, C.},
  title     = {First-Order Methods for Wasserstein Distributionally Robust MDPs},
  booktitle = {Proceedings of the International Conference on Learning Learning (ICML)},
  volume    = {139},
  pages     = {1--10},
  year      = {2021},
  doi       = {10.48550/arXiv.2009.06790}
}

@inproceedings{Yu2022,
  author    = {Yu, C. and Velu, A. and Vinitsky, E. and Gao, J. and Wang, Y. and Bayen, A. and Wu, Y.},
  title     = {{The Surprising Effectiveness of PPO in Cooperative Multi-Agent Games}},
  booktitle = {Proc. Advances in Neural Information Processing Systems (NeurIPS)},
  volume    = {35},
  pages     = {24611--24624},
  year      = {2022},
  doi       = {10.48550/arXiv.2103.01955}
}

@inproceedings{Lowe2017,
  author    = {Lowe, R. and Wu, Y. and Tamar, A. and Harb, J. and Abbeel, P. and Mordatch, I.},
  title     = {{Multi-Agent Actor-Critic for Mixed Cooperative-Competitive Environments}},
  booktitle = {Proc. Advances in Neural Information Processing Systems (NIPS)},
  pages     = {1--10},
  year      = {2017},
  doi       = {10.48550/arXiv.1706.02275}
}

@article{ShalevShwartz2012,
  author    = {Shai Shalev-Shwartz},
  title     = {Online Learning and Online Convex Optimization},
  journal   = {Foundations and Trends in Machine Learning},
  volume    = {4},
  number    = {2},
  pages     = {107--194},
  year      = {2012},
  doi       = {10.1561/2200000018}
}

@misc{NYCTLC,
  author       = {{NYC Taxi and Limousine Commission}},
  title        = {TLC Trip Record Data},
  howpublished = {\url{https://www.nyc.gov/site/tlc/about/tlc-trip-record-data.page}},
  note         = {Accessed: 2025-09-01},
  year         = {2025}
}

@misc{H3,
  author       = {{Uber Technologies, Inc.}},
  title        = {H3: A hexagonal hierarchical geospatial indexing system},
  howpublished = {\url{https://h3geo.org}},
  note         = {Accessed: 2025-09-01},
  year         = {2025}
}
\endgroup

\end{document}